\documentclass[onecolumn,11pt]{article}
\usepackage[top=1in, bottom=1in, left=1in, right=1in]{geometry}
\usepackage{amsmath,amsfonts,amscd,amssymb}
\usepackage[utf8]{inputenc} 
\usepackage[T1]{fontenc}    
\usepackage{hyperref}       
\usepackage{url}            
\usepackage{booktabs}       
\usepackage{amsfonts}       
\usepackage{nicefrac}       
\usepackage{microtype}      
\usepackage{indentfirst}
\usepackage{hyphenat}
\usepackage{amsmath}
\usepackage{amssymb}
\usepackage{mathrsfs} 
\usepackage{mathtools}

\usepackage{graphicx}
\usepackage{caption}
\usepackage{float}
\usepackage[abs]{overpic}
\usepackage{subfig}
\usepackage{epstopdf}
\usepackage{rotating}
\usepackage{multirow}
\usepackage{wrapfig}

\usepackage{enumerate}
\usepackage[numbers,sort&compress]{natbib}
\usepackage{diagbox}

\usepackage{algorithm}
\usepackage[noend]{algpseudocode}
\usepackage[utf8]{inputenc}
\usepackage{fontenc}
\usepackage{lipsum}

\usepackage[bottom,flushmargin,hang,multiple]{footmisc}
\usepackage{lipsum}
\usepackage{cancel}
\usepackage{url}
\usepackage{color}
\usepackage{mathtools}
\usepackage{subeqnarray}
\usepackage{setspace}
\usepackage{palatino} 
\usepackage{appendix}
\usepackage{soul}

\definecolor{header1}{cmyk}{0,0,0,1}

\newcommand{\norm}[1]{\left\lVert#1\right\rVert}

\newcommand{\bX}{\mathbf{X}}

\newcommand{\bsx}{\boldsymbol{x}}
\newcommand{\bsf}{\boldsymbol{f}}

\newcommand{\bXi}{\boldsymbol{\Xi}}
\newcommand{\bxi}{\boldsymbol{\xi}}

\newcommand{\bTheta}{\boldsymbol{\Theta}}

\newcommand*\R{\mathbb{R}}

\newcommand\blfootnote[1]{%
  \begingroup
  \renewcommand\thefootnote{}\footnote{#1}%
  \addtocounter{footnote}{-1}%
  \endgroup
}

\title{\vspace{-.4in}{\fontsize{16}{16}\selectfont \textbf{SINDy-PI: A Robust Algorithm for Parallel Implicit Sparse Identification of Nonlinear Dynamics}}}

\author{\normalsize{Kadierdan Kaheman$^{1*}$, J. Nathan Kutz$^2$, Steven L. Brunton$^1$}\\
\footnotesize{$^1$ Department of Mechanical Engineering, University of Washington, Seattle, WA 98195, United States}\\
\footnotesize{$^2$ Department of Applied Mathematics, University of Washington, Seattle, WA 98195, United States\vspace{-.2in}}
}
\date{}

\begin{document}
\maketitle
\blfootnote{$^*$ Corresponding author (kadierk@uw.edu); Code availalbe at github.com/dynamicslab/SINDy-PI.}

\vspace{-.2in}
\begin{abstract}
Accurately modeling the nonlinear dynamics of a system from measurement data is a challenging yet vital topic. 
The sparse identification of nonlinear dynamics~(SINDy) algorithm is one approach to discover dynamical systems models from data. 
Although extensions have been developed to identify implicit dynamics, or dynamics described by rational functions, these extensions are extremely sensitive to noise. 
In this work, we develop SINDy-PI (parallel, implicit), a robust variant of the SINDy algorithm to identify implicit dynamics and rational nonlinearities.  
The SINDy-PI framework includes multiple optimization algorithms and a principled approach to model selection. 
We demonstrate the ability of this algorithm to learn implicit ordinary and partial differential equations and conservation laws from limited and noisy data.  
In particular, we show that the proposed approach is several orders of magnitude more noise robust than previous approaches, and may be used to identify a class of ODE and PDE dynamics that were previously unattainable with SINDy, including for the double pendulum dynamics and simplified model for the Belousov–Zhabotinsky~(BZ) reaction. 
\end{abstract}

\section{Introduction}
\label{sec1}

Discovering dynamical system models from data is critically important across science and engineering.  
Traditionally, models are derived from first principles, although this approach may be prohibitively challenging in many fields, such as climate science, finance, and biology.  
Fortunately, data-driven model discovery (i.e., system identification) is a rapidly developing field~\cite{Brunton2019book}, with a range of techniques including classic linear approaches~\cite{Nelles2013book,ljung2010arc}, dynamic mode decomposition (DMD)~\cite{Schmid2010jfm,Kutz2016book} and Koopman theory~\cite{Budivsic2012chaos,Mezic2013arfm,Williams2015jnls,klus2017data}, nonlinear autoregressive models~\cite{Akaike1969annals,Billings2013book}, neural networks~\cite{yang2018physics,Wehmeyer2018jcp,Mardt2018natcomm,vlachas2018data,pathak2018model,lu2019deepxde,Raissi2019jcp,Champion2019pnas,raissi2020science}, Gaussian process regression~\cite{Raissi2017arxiva,Raissi2017arxiv}, nonlinear Laplacian spectral analysis~\cite{Giannakis2012pnas}, diffusion maps~\cite{Yair2017pnas}, genetic programming~\cite{Bongard2007pnas,Schmidt2009science,Daniels2015naturecomm}, and sparse regression~\cite{Brunton2016SINDy,Rudy2017PDE-find,Schaeffer2017prsa}, to highlight some of the recent developments.  
Of particular note is a recent push towards \emph{parsimonious} modeling~\cite{Bongard2007pnas,Schmidt2009science,Brunton2016SINDy}, which favors Pareto-optimal models with the lowest complexity required to describe the observed data.  These models benefit from being interpretable, and they tend to generalize and prevent overfitting.  
The sparse identification of nonlinear dynamics (SINDy) algorithm~\cite{Brunton2016SINDy} discovers parsimonious models through a sparsity-promoting optimization to select only a few model terms from a library of candidate functions.  
SINDy has been widely adopted in the community~\cite{Sorokina2016oe,Dam2017pf,Schaeffer2017prsa,narasingam2018data,JC2018ConGalerkin,Quade2018chaos,boninsegna2018sparse,Loiseau2018jfm,Zhang2018arxiv,Hoffmann2018arxiv,mangan2019model,Thaler2019jcp,lai2019sparse,wu2018numerical,de2019discovery,Gelss2019mindy,Guang2018}, but it relies on the dynamics having a sparse representation in a pre-defined library, making it difficult to discover implicit dynamics and rational functions. The implicit-SINDy extension~\cite{Mangan2016ImSINDY} makes it possible to identify these implicit functions, although this algorithm is extremely sensitive to noise.  
In this work, we develop a robust, parallel algorithm for the sparse identification of implicit dynamics, making it possible to explore entirely new classes of systems that were previously inaccessible.  

Parsimonious modeling has a rich history, with many scientific advances being argued on the basis of Occam's razor, that the simplest model is likely the correct one. 
SINDy exemplifies this principle, identifying a potentially nonlinear model with the fewest terms required to describe how the measurement data changes in time.  
The basic idea behind SINDy may be illustrated on a one-dimensional system $\dot{x}=f(x)$; the general formulation for multidimensional dynamics will be described in the following sections.  An interpretable form of the nonlinear dynamics may be learned by writing the rate of change of the state of the system ${x}$ as a sparse linear combination of a few terms in a library of candidate functions, $\boldsymbol{\Theta}(x)=\begin{bmatrix}\theta_1(x) & \theta_2(x) & \dotsc & \theta_p(x)\end{bmatrix}$:
\begin{align}\label{Eq:BasicSINDy}
    \dot{x}(t) = {f}({x}(t)) \approx \boldsymbol{\Theta}({x}(t))\boldsymbol{\xi}. 
\end{align}
where each $\theta_j(x)$ is prescribed candidate term (e.g. $x, x^2, \sin(x), \cdots$). 
The derivative of the state and the library of candidate functions may both be computed from measured trajectory data.   It then remains to solve for a sparse vector $\boldsymbol{\xi}$ with nonzero entries $\xi_j$ indicating which functions $\theta_j(x)$ are active in characterizing the dynamics. 
The resulting models strike a balance between accuracy and efficiency, and they are highly interpretable by construction.  
In a short time, the SINDy algorithm has been extended to include inputs and control~\cite{Kaiser2018prsa}, to identify partial differential equations~\cite{Rudy2017PDE-find,Schaeffer2017prsa}, to incorporate physically relevant constraints~\cite{JC2018ConGalerkin}, to include tensor bases~\cite{Gelss2019mindy}, and to incorporate integral terms for denoising~\cite{Schaeffer2017IntegralSINDy,Reinbold2020pre}. 
These extensions and its simple formulation in terms of a generalized linear model in \eqref{Eq:BasicSINDy} have resulted in SINDy being adopted in the fields of fluid mechanics~\cite{JC2018ConGalerkin,Loiseau2018jfm}, nonlinear optics~\cite{Sorokina2016oe}, plasma physics~\cite{Dam2017pf}, chemical reactions~\cite{Hoffmann2018arxiv,boninsegna2018sparse,narasingam2018data}, numerical methods~\cite{Thaler2019jcp}, and structural modeling~\cite{lai2019sparse}. 

The generalized linear model in \eqref{Eq:BasicSINDy} does not readily lend itself to representing implicit dynamics and rational functions, which are not naturally expressible as sum of a few basis functions.  
Instead, the implicit-SINDy algorithm~\cite{Mangan2016ImSINDY} reformulates the SINDy problem in an implicit form:
\begin{align}\label{Eq:ImplicitSimple}
    \boldsymbol{\Theta}(x,\dot{x})\boldsymbol{\xi} = 0.
\end{align}
This formulation is flexible enough to handle a much broader class of dynamics with rational function nonlinearities, such as $\dot{x} = {N(x)}/{D(x)}$ which may be rewritten as $\dot{x}D(x) + N(x) = 0$.
However, the sparsest vector $\boldsymbol{\xi}$ that satisfies \eqref{Eq:ImplicitSimple} is the trivial solution $\boldsymbol{\xi}=\mathbf{0}$. 
Thus, the implicit-SINDy algorithm leverages a recent non-convex optimization procedure~\cite{Wright2009ieeetpami,Qu2014ADM} to find the sparsest vector $\boldsymbol{\xi}$ in the null space of $\boldsymbol{\Theta}(x,\dot{x})$, which differs from other approaches~\cite{zhu2015review,zhu2018control} that identify the rational dynamics. For even small amounts of noise, the dimension of the null space will become prohibitively large, making this approach extremely sensitive to noise and compromising the model discovery process.  

This work develops an optimization and model selection framework that recasts implicit-SINDy as a convex problem, making it as noise robust as the original non-implicit SINDy algorithm and enabling the identification of implicit ODEs and PDEs that were previously inaccessible.  
The key to making the implicit-SINDy algorithm robust is the realization that if we know even a single term in the dynamics, corresponding to a non-zero entry $\xi_j$, then we can rewrite \eqref{Eq:ImplicitSimple} in a non-implicit form
\begin{align}\label{Eq:SimplePI}
    \theta_j(x,\dot{x}) = \boldsymbol{\Theta}'(x,\dot{x})\boldsymbol{\xi}'
\end{align}
where $\boldsymbol{\Theta}'$ and $\boldsymbol{\xi}'$ have the $j$-th element removed.  
Because none of these terms are known \emph{a priori}, we sweep through the  library, term by term, testing \eqref{Eq:SimplePI} for a sparse model that fits the data. 
This procedure is highly parallelizable and provides critical information for model selection.  
Our approach is related to the recent work of Zhang et al.~\cite{Guang2018}, which also makes the implicit problem more robust by testing candidate functions individually. 
However, there are a number of key differences in the present approach.  
Our work explicitly considers \emph{rational} nonlinearities to discover exceedingly complex implicit PDEs, such as a simplified model of the Belousov-Zhabotinsky (BZ) reaction. 
Our framework also provides several new greedy algorithms, including parallel and constrained formulations. 
We further extend this method to include the effect of control inputs, making it applicable to robotic systems~\cite{Koryakovskiy2018}, and we use this procedure to discover Hamiltonians.
Finally, our approach provides guidance on model selection, a comprehensive comparison with previous methods, and a careful analysis of noise robustness.

\section{Background}
\label{sec2}
We briefly introduce the full multidimensional SINDy and implicit-SINDy algorithms, which will provide a foundation for our robust implicit identification algorithm in Sec.~\ref{sec3}. 

\subsection{Sparse Identification of Nonlinear Dynamics}
The goal of SINDy~\cite{Brunton2016SINDy} is to discover a dynamical system 
\begin{align}\label{Eq:Dynamics}
    \frac{d}{dt}{\bsx}(t)=\bsf(\bsx(t)),
\end{align}
from time-series data of the state $\bsx(t)= [{x_{1}(t)},{\ldots}, {x_{n}(t)}]^{T} \in \R^{n}$. 
We assume that the dynamics, encoded by the function $\bsf$, admit a sparse representation in a library of candidate functions:
\begin{equation}
    \boldsymbol{\Theta}(\bsx) = \begin{bmatrix} \theta_1(\bsx) & \theta_2(\bsx) & \cdots & \theta_p(\bsx)\end{bmatrix}.
\end{equation}
Thus, each row equation in \eqref{Eq:Dynamics} may be written as
\begin{equation}
    \frac{d}{dt}x_k(t) = f_k(\bsx(t)) \approx \boldsymbol{\Theta}(\bsx)\boldsymbol{\xi}_k
\end{equation}
where $\boldsymbol{\xi}_k$ is a sparse vector, indicating which terms are active in the dynamics.  

We determine the nonzero entries of $\boldsymbol{\xi}_k$ through sparse regression based on trajectory data. 
The time-series data is arranged into a matrix ${\bX = \begin{bmatrix} \bsx(t_1) & \bsx(t_2) & \cdots & \bsx(t_m)\end{bmatrix}^T}$, and the associated time derivative matrix ${\dot{\bX}=\begin{bmatrix}\dot{\bsx}(t_1)& \dot{\bsx}(t_2)& \cdots & \dot{\bsx}(t_m)\end{bmatrix}^T}$ is computed using an appropriate numerical differentiation scheme~\cite{Chartrand2011TVReg,Brunton2016SINDy,Brunton2019book}. 
It is then possible to evaluate the library $\boldsymbol{\Theta}$ on trajectory data in $\bX$ so that each column of $\boldsymbol{\Theta}(\bX)$ is a function $\theta_j$ evaluated on the $m$ snapshots in $\bX$. 

It is now possible to write the dynamical system in terms of a generalized linear model, evaluated on trajectory data:
\begin{equation}\label{Eq:SINDyMatrix}
    \dot{\bX}=\boldsymbol{\Theta}(\bX) \boldsymbol{\Xi}.
\end{equation}
There are several approaches to identify the sparse matrix of coefficients $\boldsymbol{\Xi}$, including sequentially thresholded least squares~(STLSQ)~\cite{Brunton2016SINDy,zhang2019convergence}, LASSO~\cite{Tibshirani1994}, sparse relaxed regularized regression~(SR3)~\cite{ZhengSR3,Champion2019SINDySR3}, stepwise sparse regression~(SSR)~\cite{boninsegna2018sparse}, and Bayesian approaches~\cite{Guang2018,Pan2016BayesianSINDy}. 
It is possible to augment the library to include partial derivatives for the identification of partial differential equations (PDEs)~\cite{Rudy2017PDE-find,Schaeffer2017prsa}.  
Similarly, it is possible to include external forcing terms in the library $\boldsymbol{\Theta}$, enabling the identification of forced and actively controlled systems~\cite{Kaiser2018prsa}. To alleviate the effect of noise, it is possible to reframe the SINDy problem in terms of an integral formulation~\cite{Schaeffer2017IntegralSINDy,Reinbold2020pre}. There are a number of factors that affect the robustness of SINDy, some of which are discussed in App.~\ref{App:Robustness}.

\subsection{Implicit Sparse Identification of Nonlinear Dynamics}
The implicit-SINDy algorithm~\cite{Mangan2016ImSINDY} extends SINDy to identify implicit differential equations
\begin{equation}\label{Eq:ImplicitDynamics}
    \bsf(\bsx,\dot{\bsx})=0,
\end{equation}
and in particular, systems that include rational functions in the dynamics, such as chemical reactions and metabolic networks that have a separation of timescales.  

The implicit-SINDy generalizes the library $\boldsymbol{\Theta}(\bX)$ in \eqref{Eq:SINDyMatrix} to include functions of $\bsx$ and $\dot{\bsx}$:
\begin{equation}\label{Eq:implicitSINDyMatrix}
    \boldsymbol{\Theta}(\bX,\dot{\bX}) \boldsymbol{\Xi}=\mathbf{0}. 
\end{equation}
However, this approach requires solving for a matrix $\boldsymbol{\Xi}$ whose columns $\boldsymbol{\xi}_k$ are sparse vectors in the null space of $\boldsymbol{\Theta}(\bX,\dot{\bX})$.  
This approach is non-convex, relying on the alternating directions method (ADM)~\cite{Mangan2016ImSINDY,Qu2014ADM}, and null space computations are highly ill-conditioned for noisy data~\cite{Gavish2014ieeetit,Mangan2016ImSINDY,Brunton2019book}, thus inspiring the current work and mathematical innovations.

\begin{figure}[t]
\vspace{-.2in}
    \centering
    \includegraphics[width=1\textwidth]{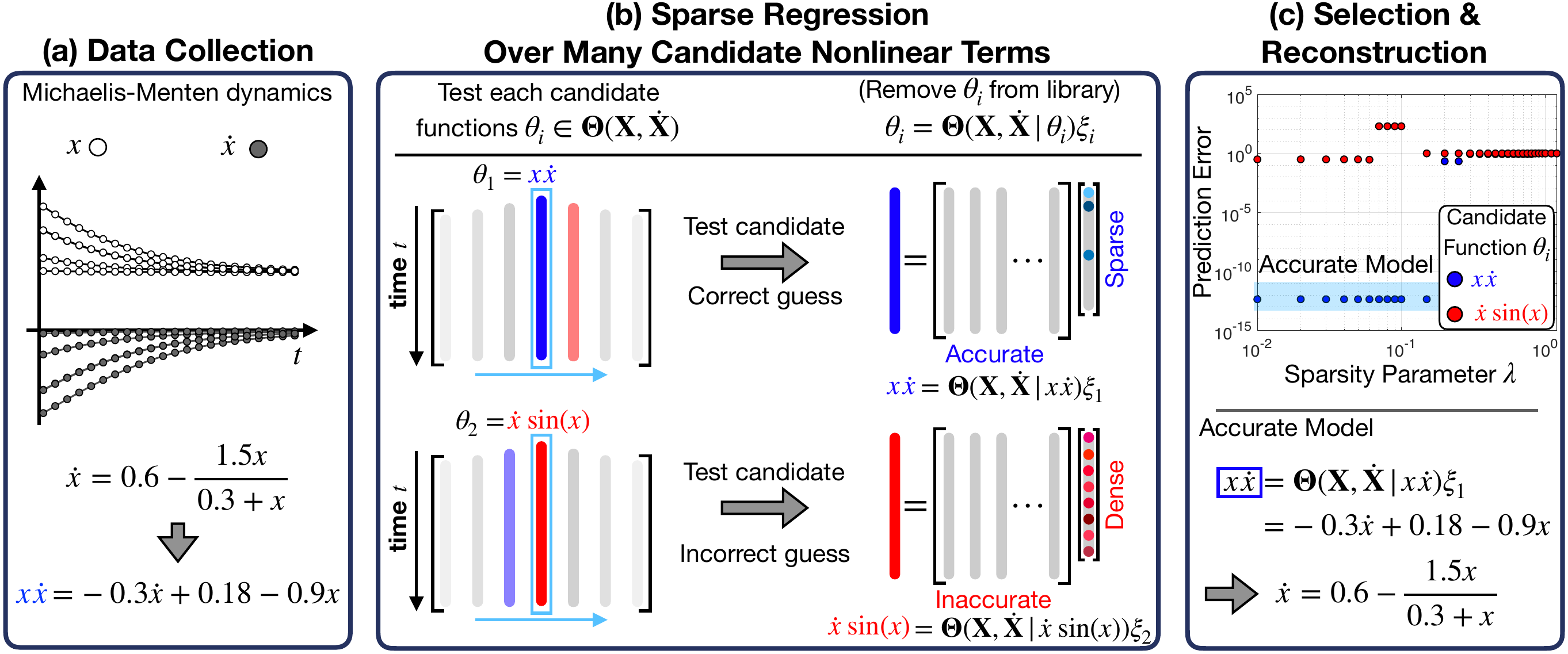}
    \caption{The illustration of the SINDy-PI algorithm on Michaelis-Menten dynamics. (a) The Michaelis-Menten system is simulated, and measurement data is provided to SINDy-PI. (b) Multiple possible left-hand side functions are tested at the same time. (c) The candidate model prediction error is calculated, and the best model is selected.}
    \label{Fig_DL_SINDy}
\end{figure}
\section{SINDy-PI: Robust Parallel Identification of Implicit Dynamics}\label{sec3}
We have developed the SINDy-PI (parallel, implicit) framework for the robust identification of implicit dynamics, bypassing the null space approach discussed in Sec.~2.2.  
The idea is that if even a single term $\theta_j(\bsx,\dot{\bsx})\in\boldsymbol{\Theta}(\bsx,\dot{\bsx})$ in the dynamics \eqref{Eq:ImplicitDynamics} is known, it is possible to rewrite \eqref{Eq:implicitSINDyMatrix} as
\begin{equation}\label{Eq:SINDyPIMatrix}
    \theta_j(\bX,\dot{\bX}) = \boldsymbol{\Theta}(\bX,\dot{\bX}|\theta_j(\bX,\dot{\bX}) \boldsymbol{\xi}_j,
\end{equation}
where $\boldsymbol{\Theta}(\bX,\dot{\bX}|\theta_j(\bX,\dot{\bX}))$ is the library $\boldsymbol{\Theta}(\bX,\dot{\bX})$ with the $\theta_j$ column removed. 
Equation~\eqref{Eq:SINDyPIMatrix} is no longer in implicit form, and the sparse coefficient matrix corresponding to the remaining terms may be solved for using previously developed SINDy techniques~\cite{Brunton2016SINDy,ZhengSR3,Champion2019SINDySR3,boninsegna2018sparse,Guang2018,Pan2016BayesianSINDy,Rudy2017PDE-find,Schaeffer2017prsa,Schaeffer2017IntegralSINDy,Reinbold2020pre}. In particular, we solve for a sparse coefficient vector $\boldsymbol{\xi}_j$ that minimizes the following loss function:
\begin{equation}\label{Eq:SINDyPILoss}
     \| \theta_j(\bX,\dot{\bX}) - \boldsymbol{\Theta}(\bX,\dot{\bX}|\theta_j(\bX,\dot{\bX}) \boldsymbol{\xi}_j\|_2 +\beta\norm{\boldsymbol{\xi}_j}_0,
\end{equation}
where $\beta$ is the sparsity promoting parameter.
There are numerous relaxations of the non-convex optimization problem in~\eqref{Eq:SINDyPILoss}, for example the sequentially thresholded least-squares algorithm~\cite{Brunton2016SINDy}. Because there is no null space calculation, the resulting algorithm is considerably more robust to noise than the implicit-SINDy algorithm~\cite{Mangan2016ImSINDY}, i.e. we longer have to deal with an ill-conditioned null space problem. 

In general, the entire point of SINDy is that the dynamics are not known ahead of time, and so it is necessary to test each candidate function $\theta_j$ until one of the models in \eqref{Eq:SINDyPIMatrix} admits a sparse and accurate solution.  
When an incorrect candidate term is used, then the algorithm results in a dense (non-sparse) model $\boldsymbol{\xi}_j$ and an inaccurate model fit, and when a correct term is included, the algorithm identifies a sparse model $\boldsymbol{\xi}_j$ and an accurate model fit.  
In this way, it is clear when the algorithm has identified the correct model.  
Moreover, there is a wealth of redundant information, since each term in the correct model may be used as the candidate function on the left hand side, and the resulting models may be cross-referenced. 
This approach is highly parallelizable, and each candidate term may be tested simultaneously in parallel.  
The non-parallel formulation in \eqref{Eq:SINDyPIMatrix} was recently introduced by Zhang et al.~\cite{Guang2018} in the context of Bayesian regression, where they also make the implicit problem more robust by testing candidate functions individually; however, they do not consider dynamics with rational function nonlinearities or control inputs. 
In this work, we extend the robust implicit formulation to identify several challenging implicit ODE and PDE systems with rational function nonlinearities, which are ubiquitous in engineering and natural systems, and systems with external forcing and control inputs.  
We also introduce the parallel formulation and model selection frameworks.  
Further, we will introduce a constrained optimization framework to simultaneously test all candidate functions.

\subsection{Model Selection}
For each candidate function in \eqref{Eq:SINDyPIMatrix}, we obtain one candidate model.  
When the candidate function $\theta_j$ is not in the true dynamics, then the resulting coefficient vector $\boldsymbol{\xi}_j$ will not be sparse and there will be large prediction error.  
In contrast, when a correct candidate function is selected, then we obtain a sparse coefficient vector $\boldsymbol{\xi}_j$ and small prediction error.  
For an implicit dynamical system, there may be several different implicit equations that must be identified, resulting in several candidate functions that admit sparse models.  
The sequentially thresholded least squares (STLSQ) algorithm that we use here, and whose convergence properties are considered by Zhang and Schaeffer~\cite{zhang2019convergence}, iteratively computes a least-squares solution to minimize $\|\theta_j(\bX,\dot{\bX})-\boldsymbol{\Theta}(\bX,\dot{\bX}|\theta_j(\bX,\dot{\bX}))\boldsymbol{\xi}_j\|_2$ and then zeros out small entries in $\boldsymbol{\xi}_j$ that are below a set threshold $\lambda$.  
This threshold $\lambda$ is a hyperparameter that must be tuned to select the model that most accurately balances accuracy and efficiency.  
Thus, we must employ model selection techniques to identify the implicit models that best supports the data, while remaining as simple as possible.  

There are several valid approaches to model selection.  To select a parsimonious yet accurate model we can also employ the Akaike information criterion~(AIC)~\cite{Akaike1998InformationTheory, Akaike1974StatisticalModelIdentification} and Bayesian information criterion~(BIC)~\cite{Schwarz1978EstimatingDomensionOfAModel}, as in~\cite{Mangan2017ModelSelection}. 
It is also possible to sweep through the parameter $\lambda$ and candidate functions $\theta_j$, and then choose the Pareto optimal model from a \emph{family} of models on the Pareto front balancing accuracy and efficiency; this is the approach in the original SINDy work~\cite{Brunton2016SINDy} and in earlier work leveraging genetic programming to discover dynamics~\cite{Bongard2007pnas,Schmidt2009science}.  
In this work, we take a different approach, selecting models based on performance on a test data set $\bX_t$ that has been withheld for model validation to automate the model selection process.  For each threshold $\lambda$, the resulting model is validated on the test set $\bX_t$, and the model with the lowest test error is selected.  
One error function is the model fit:
\begin{equation}
    \label{eq27}
    \text{Error}={\frac{\norm{\theta_j(\bX_t,\dot{\bX}_t)-\bTheta(\bX_t,\dot{\bX}_t|\theta_j(\bX_t,\dot{\bX}_t)) \bXi}_2}{\norm{\theta_j(\bX_t,\dot{\bX}_t)}_2}}.
\end{equation}
In practice, for rational dynamics, we select based upon the predicted derivative $\dot{\bX}_t$:
\begin{equation}
    \label{eq_n1}
    \text{Error}={\frac{\norm{\dot{\bX}_t-\dot{\bX}^{\text{model}}_t}_2}{\norm{\dot{\bX}_t}_2}}.
\end{equation}
For implicit dynamics where each state derivative may be written as a rational function 
\begin{equation}
    \dot{x}_k = f_k(\bsx) = \frac{N_k(\bsx)}{D_k(\bsx)},
\end{equation}
then we restrict the candidate functions to $\theta_j(\bsx,\dot{\bsx})=\dot{x}_k\theta_j(\bsx)$ for some $\theta_j(\bsx)\in\boldsymbol{\Theta}(\bsx)$ to identify a separate sparse model for each $\dot{x}_k$.  Several candidate functions may provide accurate and sparse models.  These different models may further be cross-references to check that the same terms are being selected in each model, providing additional information for model selection and validation.  

\subsection{Constrained Optimization Formulation}
\label{sec3.2}

\begin{figure}[t]
\vspace{-.2in}
    \centering
    \includegraphics[width=1\textwidth]{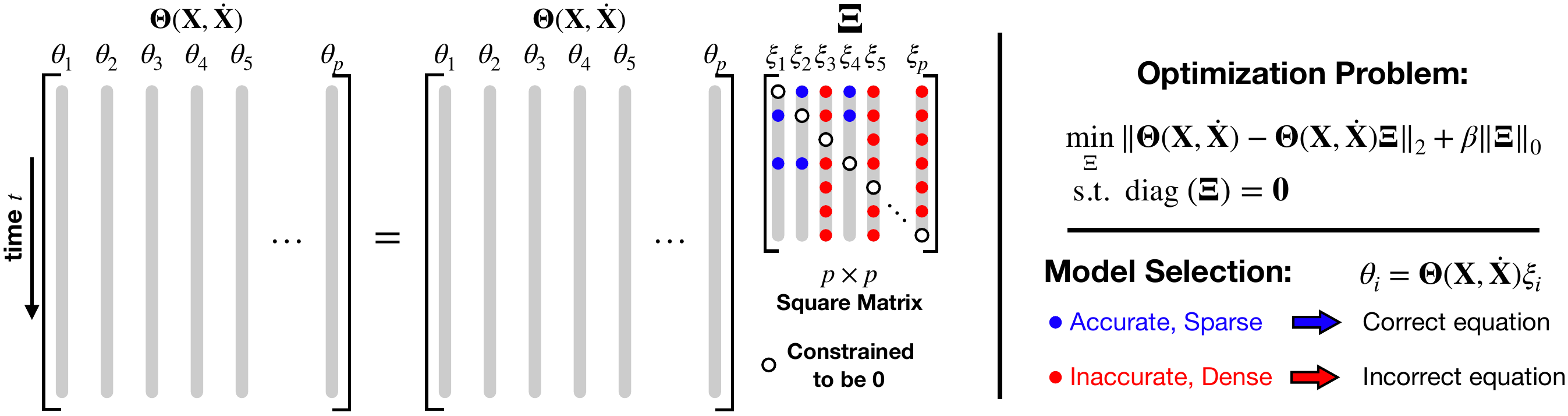}
    \vspace{-.2in}
    \caption{Schematic illustrating the constrained formulation of the SINDy-PI algorithm. }
    \label{Fig_DL_SINDy_Con}
\end{figure}

In \eqref{Eq:SINDyPIMatrix} each candidate function was tested individually in a parallel optimization.  
However, each of these individual equations may be combined into a single constrained system of equations
\begin{equation}
    \label{eq43}
    \bTheta(\bX,\dot{\bX})=\bTheta(\bX,\dot{\bX})\bXi \quad\text{such that}\quad \bXi_{jj}=0.
\end{equation}
We constrain $\boldsymbol{\Xi}$ to have zero entries on the diagonal, as shown in Fig.~\ref{Fig_DL_SINDy_Con}, which is the same as removing the candidate function from the library in the separate optimization problems in \eqref{Eq:SINDyPIMatrix}.  
Without this constraint, the trivial solution $\boldsymbol{\Xi}=\mathbb{I}_{p\times p}$ will provide the sparsest $\boldsymbol{\Xi}$ and the most accurate model. 
This may be written as a formal constrained optimization problem:
\begin{equation}
    \label{eq45}
    \begin{aligned}
        \min_{\bXi} \quad & ||\bTheta(\bX,\dot{\bX})-\bTheta(\bX,\dot{\bX})\bXi||_2 + \beta\|\bXi\|_0,\\
        \textrm{s.t.} \quad & \textrm{diag}(\bXi)=\mathbf{0}.
    \end{aligned}
\end{equation}
This optimization is non-convex, although there are many relaxations that result in accurate and efficient proxy solutions~\cite{Tibshirani1994,Brunton2016SINDy,ZhengSR3}. 
In this work, we will use sequentially thresholded least squares, so that any entry $\bXi_{ij}<\lambda$ will be set to zero; the sparsity parameter $\lambda$ is a hyperparameter, and each column equation may require a different parameter $\lambda_j$.  
The constrained formulation in \eqref{eq45} can be solved efficiently in modern optimization packages, and we use CVX~\cite{CVX_Software,Grant_Boyd2008_Nonsmooth_Convex_Programs}. 
After solving \eqref{eq45} we have numerous candidate models, one for each column $\bxi_k$ of $\bXi$, given by
\begin{equation}
    \label{eq46}
    \bTheta(\bX,\dot{\bX})\bxi_j = 0.
\end{equation}
The sparse models that result in an accurate fit are candidate implicit models, and they may be assessed using the model selection approaches outlined above. 
These various models may be cross-referenced for consistency, as the same models will have the same sparsity pattern.  
This information can then be used to refine the library $\bTheta$, for example to only include the nonzero entries in the sparse columns of $\bXi$.  

\subsection{Noise Robustness}
\label{sec3.3}
\begin{figure}[t]
    \vspace{-.2in}
    \centering
    \includegraphics[width=1\textwidth]{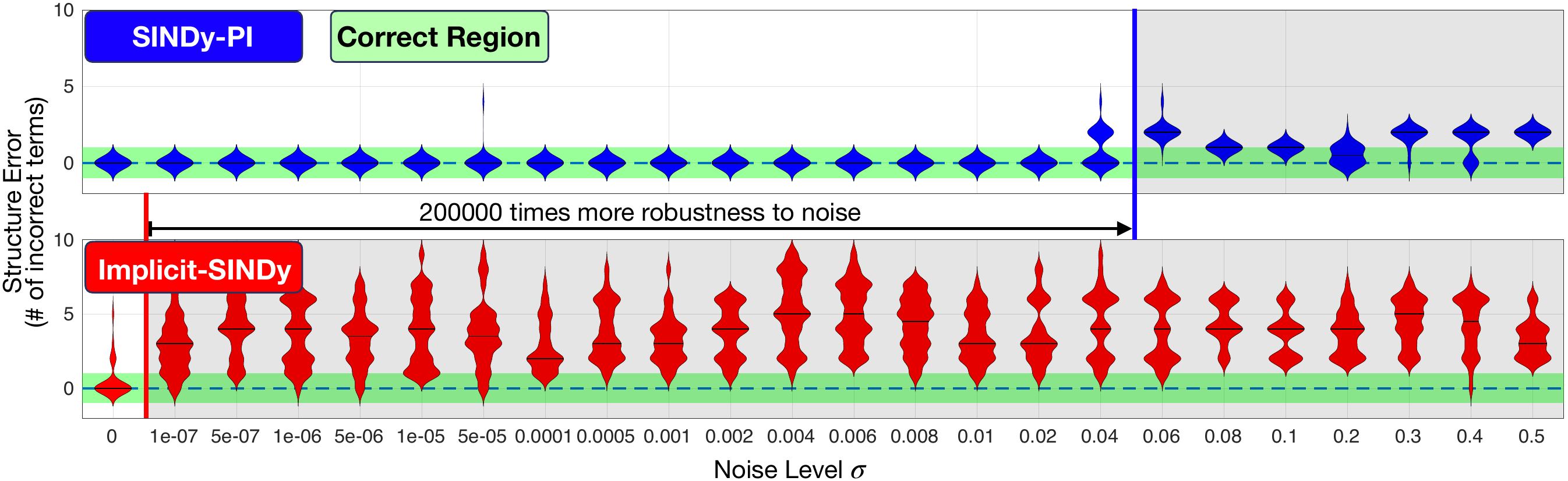}
    \caption{SINDy-PI and implicit-SINDy are compared on the Michaelis-Menten kinetics, where the structure error quantifies the number of terms in the model that are incorrectly added or deleted, compared with the true model. The derivative is computed by the total-variation regularization difference~(TVRegDiff)~\cite{Chartrand2011TVReg} on noisy state measurements. The violin plots show the cross-validated distribution of the number of incorrect terms across $30$ models. The green region indicates no structural difference between the identified model and the ground truth model. Details are provided in Appendix~\ref{App:NumericalExperiments}.}
    \label{TVReg_k0}
\end{figure}

We now compare the noise sensitivity of SINDy-PI and implicit-SINDy on the one-dimensional Michaelis–Menten model for enzyme  kinetics~\cite{Johnson2011MMK_Translate,Mangan2016ImSINDY,Cornish_Bowden2015MMK}, given by
\begin{equation}
    \label{s4eq1}
    \dot{x}=j_{x}-\frac{V_{\max } x}{K_{m}+x},
\end{equation}
where $x$ denotes the concentration of the substrate, $j_x$ denotes the influx of the substrate, $V_{max}$ denotes the maximum reaction time, and $K_m$ represents the concentration of half-maximal reaction. 
We use the same parameters as in~\cite{Mangan2016ImSINDY}, with $j_x=0.6$, $V_{max}=1.5$, and $K_m=0.3$. 
Figure~\ref{TVReg_k0} shows the result of the noise robustness of SINDy-PI and implicit-SINDy. 
In this example, SINDy-PI is able to handle over $10^5$ more measurement noise than implicit-SINDy, while still accurately recovering the correct model.  Details are provided in App.~\ref{A1}, and key factors that limit robustness are discussed inn App.~\ref{App:Robustness}.

\subsection{Data Usage}
\label{sec3.4}
\begin{figure}[t]
    \vspace{-.05in}
    \centering
    \includegraphics[width=1\textwidth]{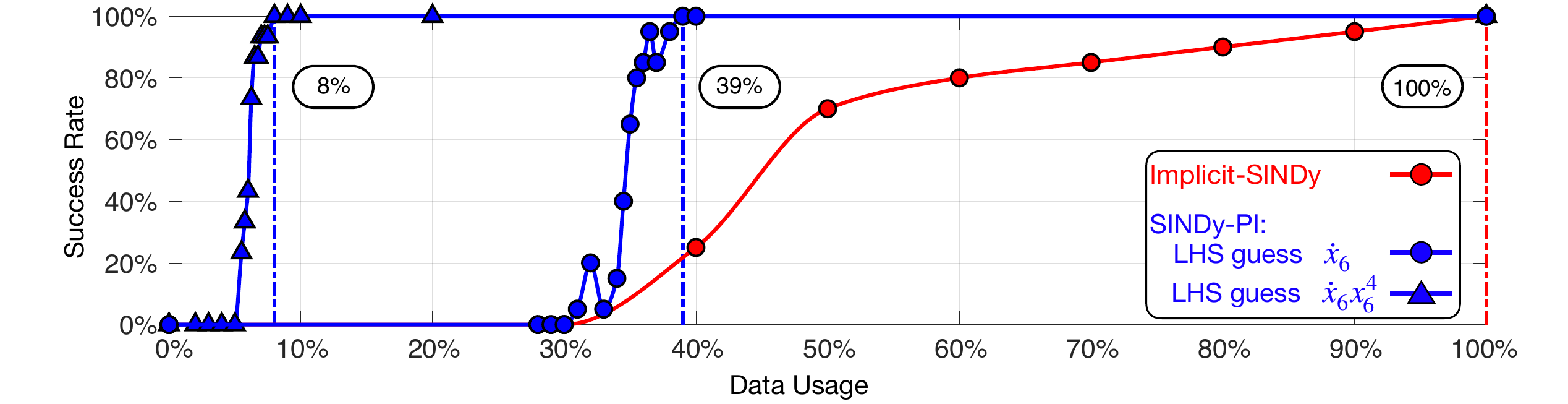}
    \caption{Success rate of SINDy-PI and implicit-SINDy identifying yeast glycolysis~\eqref{s4eq4sub6} with different percentage of training data. Each data usage percentage is randomly sampled from the entire data set composed of all trajectories.  The success rate is calculated by averaging the results of $20$ runs.}
    \label{Data_Length_Compare}
\end{figure}

The data required to correctly identify a model is a critical aspect when comparing SINDy-PI and implicit-SINDy. 
Many experimental data sets are limited in volume, and thus our goal is to identify a model with as little data as possible.  In this section, we compare the SINDy-PI and implicit-SINDy methods on the challenging yeast glycolysis model~\cite{Schmidt2011YeastGlycolysis,Mangan2016ImSINDY} given by
\begin{subequations}
    \label{s4eq4}
    \begin{align}
       \dot{x}_{1} & =c_{1}+\frac{c_{2} x_{1} x_{6}}{1+c_{3} x_{6}^{4}}\label{s4eq4sub1},\\ 
       \dot{x}_{2} & =\frac{d_{1} x_{1} x_{6}}{1+d_{2} x_{6}^{4}}+d_{3} x_{2}-d_{4} x_{2} x_{7}\label{s4eq4sub2},\\ 
       \dot{x}_{3} & =e_{1} x_{2}+e_{2} x_{3}+e_{3} x_{2} x_{7}+e_{4} x_{3} x_{6}+f_{5} x_{4} x_{7}\label{s4eq4sub3},\\ 
       \dot{x}_{4} & =f_{1} x_{3}+e_{2} x_{4}+f_{3} x_{5}+f_{4} x_{3} x_{6}+f_{5} x_{4} x_{7}\label{s4eq4sub4},\\ 
       \dot{x}_{5} & =g_{1} x_{1}+g_{2} x_{5}\label{s4eq4sub5},\\ 
       \dot{x}_{6} & =h_{3} x_{3}+h_{5} x_{6}+h_{4} x_{3} x_{6}+\frac{h_{1} x_{1} x_{6}}{1+h_{2} x_{6}^{4}}\label{s4eq4sub6},
       \\ \dot{x}_{7} & =j_{1} x_{2}+j_{2} x_{2} x_{7}+j_{3} x_{4} x_{7}\label{s4eq4sub7}.  
    \end{align}
\end{subequations}
 Equation~\eqref{s4eq4sub6} is the most challenging equation to discover in this system, and Fig.~\ref{Data_Length_Compare} compares the success rate of SINDy-PI and implicit-SINDy in identifying this equation.  SINDy-PI uses about $12$ times less data than the implicit-SINDy when identifying~\eqref{s4eq4sub6}. Details are provided in Appendices~\ref{B1} and~\ref{E1}.

\subsection{Comparison for Implicit PDE Identification}
\label{sec3.5}

\begin{figure}[t]
    \centering
    \includegraphics[width=1\textwidth]{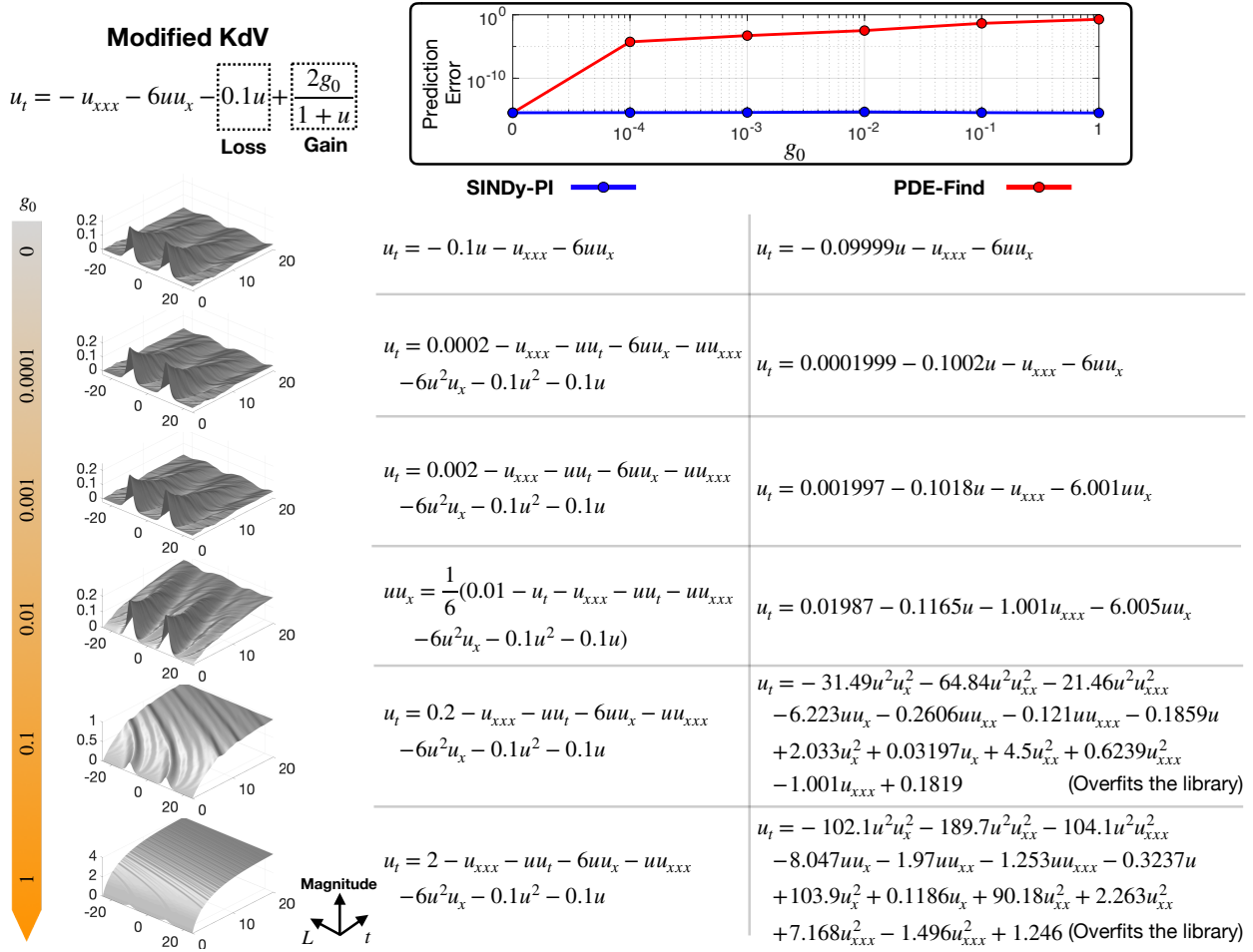}
    \caption{Comparison of SINDy-PI and PDE-FIND on an implicit PDE problem given by the modified KdV equation \eqref{s5.3.1eq2}. As we increase $g_0$, the rational term begins to play a significant role in the system behavior. For small $g_0$, PDE-FIND compensates for the effect of the rational term by tuning the other coefficients. When $g_0$ is  large, PDE-FIND overfits the library. SINDy-PI, on the other hand, correctly identifies the rational term.}
    \label{PDE_KdV}
\end{figure}

We now investigate the ability of SINDy-PI to discover a PDE with rational terms, given by a modified KdV equation 
\begin{equation}
    \label{s5.3.1eq2}
    u_t=-u_{xxx}-6uu_{x}-\gamma u+\frac{2g_0}{1+u},
\end{equation}
where $\gamma u$ is a loss term and $2g_0/(1+u)$ is a gain term.  
We fix $\gamma=0.1$ and vary the value of $g_0$ from $0$ to $1$. 
As $g_0$ increases, the implicit term gradually dominates the dynamics. 
Figure~\ref{PDE_KdV} shows the results of SINDy-PI and PDE-FIND~\cite{Rudy2017PDE-find} for different values of $g_0$. 
For large $g_0$, SINDy-PI is able to accurately identify the rational function term, while this is not possible for PDE-FIND, since this term is not in the library.  
Details of the identification process are given in App.~\ref{D1}.

\section{Advanced Examples}
\label{sec5}
We will now demonstrate the SINDy-PI framework on several challenging examples, including the double pendulum, an actuated single pendulum on a cart, the Belousov-Zhabotinsky PDE, and the identification of conserved quantities.  All examples are characterized by rational nonlinearities, and we were unable to identify them using SINDy or implicit-SINDy, even in the absence of noise.  

\subsection{Mounted Double Pendulum}
\label{secDoublePen}

\begin{figure}[t]
\vspace{-.25in}
    \centering
    \includegraphics[width=1\textwidth]{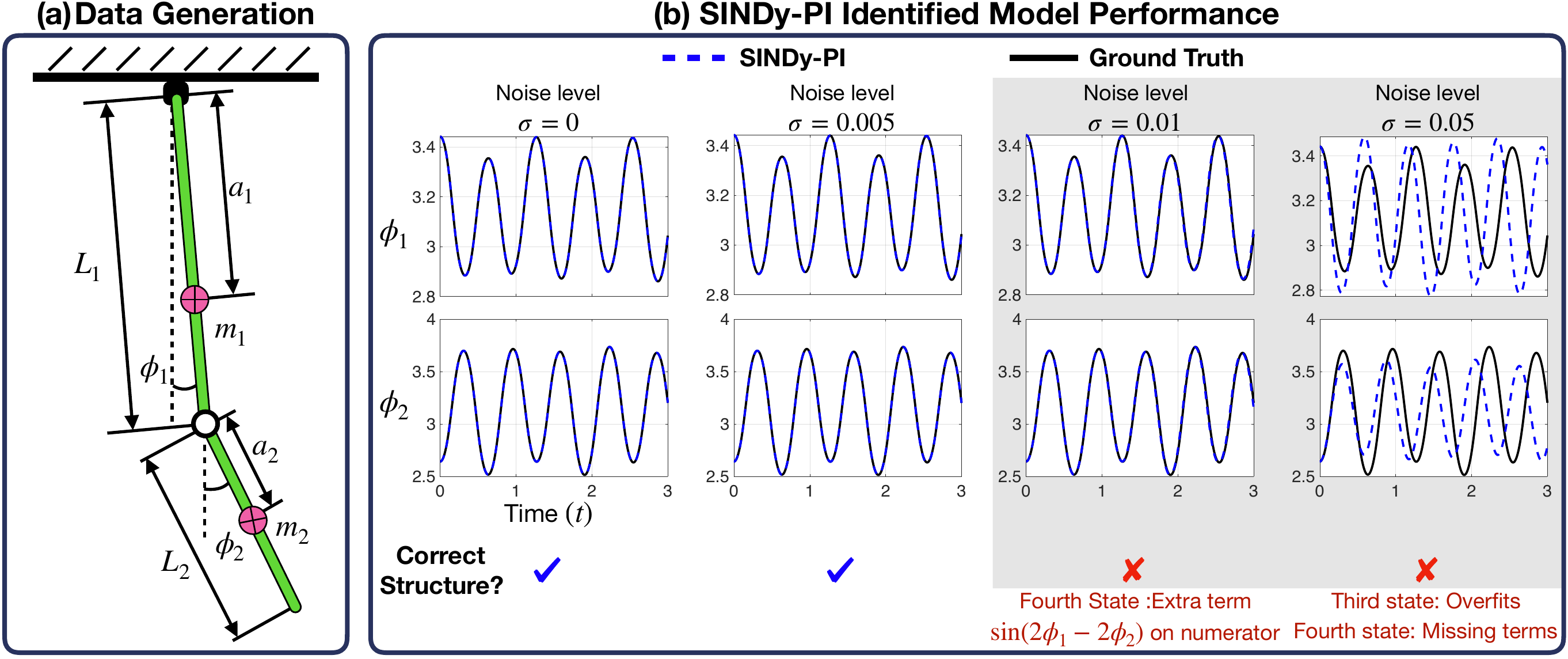}
    \vspace{-.15in}
    \caption{Schematic illustration of SINDy-PI identifying a mounted double pendulum system.}
    \label{DoublePenFig}
    \vspace{-.075in}
\end{figure} 

In our first example, we use SINDy-PI to discover the equations of motion of a mounted double pendulum, shown in Fig.~\ref{DoublePenFig}. 
The double pendulum is a classic example of chaotic dynamics~\cite{Graichen2007}, and was an original challenging example used to demonstrate the capabilities of genetic program for model discovery~\cite{Schmidt2009science}. 
Correctly modeling the nonlinear dynamics is vital for accurate control~\cite{Graichen2007}. 

We simulate the double pendulum dynamics, derived from the Euler-Lagrange equations, and use SINDy-PI to re-discover the dynamics from noisy measurements of the trajectory data. 
The governing equations and SINDy-PI models are provided in App.~\ref{App:SINDyDoublePendulum}. 
Because these dynamics have rational nonlinearities, the original SINDy algorithm is unable to identify the dynamics, making this a challenging test case. 
The state vector is given by $\bsx= [\phi_1, \phi_2,  \dot{\phi}_1, \dot{\phi}_2]^T$, and the parameters of the simulation are given in App.~\ref{E1}.  The training data is generated from an initial condition $x_{\text{train}}=[\pi+1.2,\ \pi-0.6,\ 0,\ 0]^T$, simulated for $10$ seconds using a time step of $dt=0.001$ seconds. The validation data is generated from an initial condition $x_{\text{val}}=[\pi-1,\ \pi-0.4,\ 0.3,\ 0.4]^T$, simulated  for $3$ seconds with time step $dt=0.001$ seconds.


To test the robustness of SINDy-PI, we add Gaussian noise to both the training and validation data.  We test the resulting models using a new testing initial condition $x_{\text{test}}=[\pi+0.3,\ \pi-0.5,\ 0,\ 0]^T$. 
We construct our library $\boldsymbol{\Theta}$ to include over 40 trigonometric and polynomial terms. 
The most challenging part of this example is building a library with the necessary terms, without it growing too large. The library cannot be too extensive, or else the matrix $\boldsymbol{\Theta}$ becomes ill conditioned, making it sensitive to noise. 
To reduce the library size, we use one piece of expert knowledge: the trigonometric terms should only consist of $\phi_1$ and $\phi_2$, the rotational angles of the pendula. 

The candidate functions are chosen as a combination of state derivatives and trigonometric functions. Fig.~\ref{DoublePenFig} shows that SINDy-PI can identify the equations of motion for low noise.  For larger noise, SINDy-PI misidentifies the dynamics, although it still has short term prediction ability. 


\subsection{Single Pendulum on a Cart}
\label{secSinglePen}
\begin{figure}[t]
    \vspace{-.25in}
    \centering
    \includegraphics[width=1\textwidth]{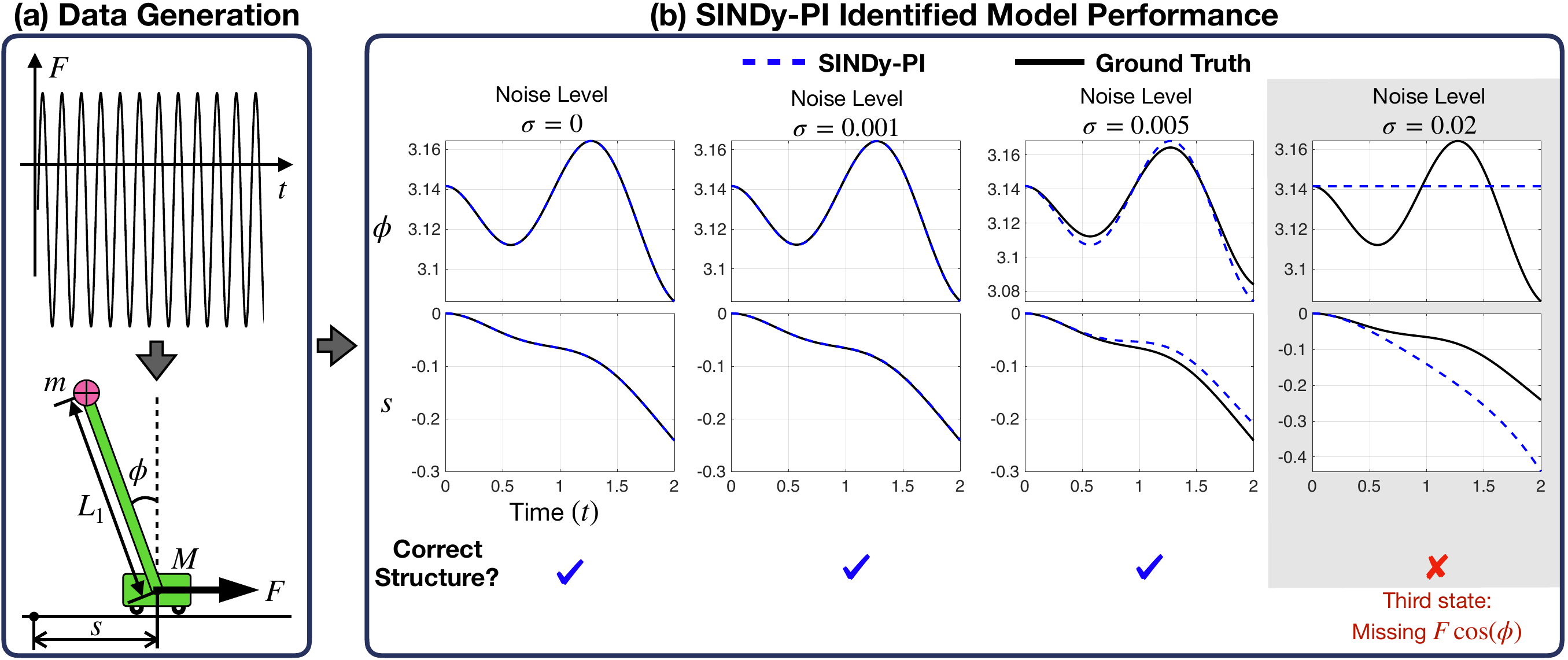}
    \vspace{-.175in}
    \caption{SINDy-PI is used to identify the single pendulum on a cart system. Control  is applied to the cart, and both the cart and pendulum states are measured. When the measurement noise is small, SINDy-PI can identify the correct structure of the model. }
    \vspace{-.075in}
    \label{Single_Pen}
\end{figure}

We now apply SINDy-PI to identify a fractional ODE problem with control input, given by the single pendulum on a cart in Fig.~\ref{Single_Pen}. SINDy has already been extended to include control inputs~\cite{Kaiser2018prsa}, although the original formulation doesn't accommodate rational functions.  

The dynamics are derived from the Euler-Lagrange equations. All system parameters except for gravity are chosen to be $1$, as summarized in App.~\ref{E1}; the governing equations and SINDy-PI models are shown in App.~\ref{App:SINDyPendulumCart}.  
The cart position is denoted by $s$. 
The state vector is given by $\bsx= [\phi, s,  \dot{\phi}, \dot{s}]^T$. 
The equations of motion are given by
\begin{subequations}
    \label{s5.2eq2}
    \begin{align}
        \frac{d}{dt}\phi&=\dot{\phi},\\
        \frac{d}{dt}s&=\dot{s},\\
        \frac{d}{dt}\dot{\phi}&=-\frac{(M+m)g\sin{(\phi)}+FL_1\cos{(\phi)}+mL_1^2\sin{(\phi)}\cos{(\phi)}\dot{\phi}^2}{L_1^2(M+m-m\cos{(\phi)}^2)},\\
        \frac{d}{dt}\dot{s}&=\frac{mL_1^2\sin{(\phi)}\dot{\phi}^2+FL_1+mg\sin{(\phi)}\cos{(\phi)}}{L_1(M+m-m\cos{(\phi)}^2)},
    \end{align}
\end{subequations}


Eq.~\eqref{s5.2eq2} is simulated with a time step of $dt=0.001$ to generate the training and testing data for model selection. The training data is generated using an initial condition $x_{\text{train}}=[0.3,\ 0,\ 1,\ 0]^T$ with the control input chosen as $F_{\text{train}}=-0.2+0.5\sin{(6t)}$, for time $t=0$ to $t=16$. 
Similarly, the validation data is generated using an initial condition $x_{\text{val}}=[0.1,\ 0,\ 0.1,\ 0]^T$ with the control input chosen as $F_{\text{val}}=-1+\sin{(t)}+3\sin{(2t)}$, for time  $t=0$ to $t=2$. 

 The library is constructed using a combination of trigonometric and polynomial terms. Around $50$ different basis functions are used for the library, and around $10$ terms are tested as candidate functions. We add Gaussian noise to all system states. We then test the SINDy-PI model on a testing initial condition $x_{\text{test}}=[\pi,\ 0,\ 0,\ 0]^T$ with control input $F_{\text{test}}=-0.5+0.2\sin{(t)}+0.3\sin{(2t)}$ for time $t=0$ to $t=2$. 
 Fig.~\ref{Single_Pen} shows the resulting SINDy-PI models. The structure of the model is correctly identified up to a noise magnitude of $0.01$. Beyond this noise level, the SINDy-PI identified model only has short term prediction ability. 
\subsection{Simplified Model of the Belousov–Zhabotinsky Reaction}
\label{secBZ}
\begin{figure}[t]
    \vspace{-.2in}
    \centering
    \includegraphics[width=1\textwidth]{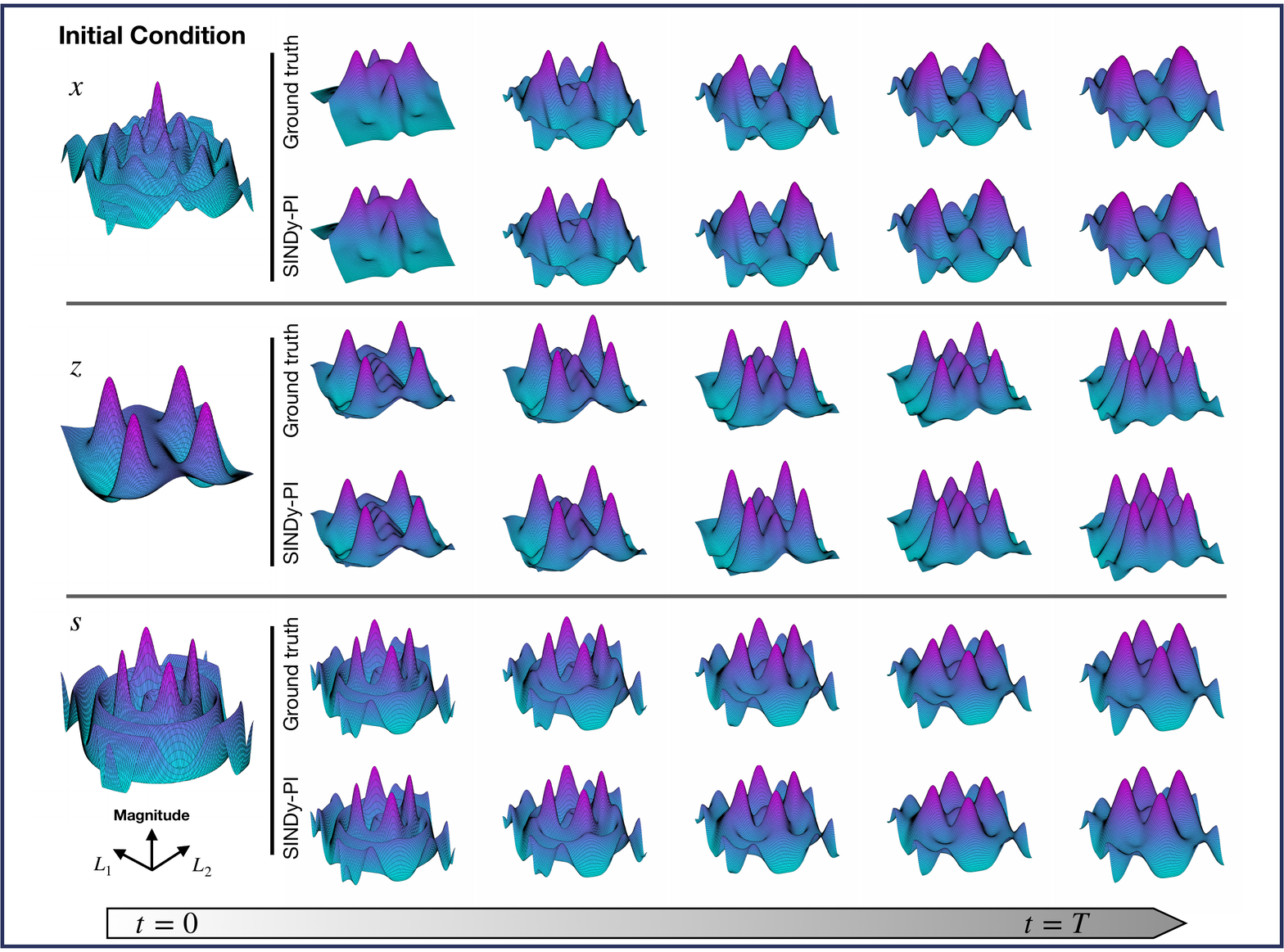}
    \caption{SINDy-PI is able to identify the simplified Belousov–Zhabotinsky reaction model.}
    \label{BZ_Reaction}
\end{figure}

We now apply SINDy-PI to a challenging PDE with rational nonlinearities, a simplified model of the Belousov\hyp{}Zhabotinsky~(BZ) reaction. The simplified BZ reaction model is given by~\cite{Vanag2004BZ_Reaction}
\begin{subequations}
    \label{s5.3.2eq1}
    \begin{align}
       \frac{\partial x}{\partial \tau}&=\frac{1}{\varepsilon}\left(\frac{fz(q-x)}{q+x}+x-x^2-\beta x+s\right)+\frac{D_x}{D_u}\Delta x,
       \label{s5.3.2eq1sub1}
       \\
       \frac{\partial z}{\partial \tau}&=x-z-\alpha z+\gamma u+\frac{D_z}{D_u}\Delta z,
       \label{s5.3.2eq1sub2}
       \\
       \frac{\partial s}{\partial \tau}&=\frac{1}{\varepsilon_2}(\beta x-s+\chi u)+\frac{D_s}{D_u}\Delta s,
       \label{s5.3.2eq1sub3}
       \\
       \frac{\partial u}{\partial \tau}&=\frac{1}{\varepsilon_3}[\alpha z -(\gamma +\frac{\chi}{2})u]+\frac{D_u}{D_u}\Delta u,
       \label{s5.3.2eq1sub4}
    \end{align}
\end{subequations}
where $x$, $z$, $s$, and $u$ are dimensionless variables and $\Delta=\frac{\partial^2}{\partial x_s^2}+\frac{\partial^2}{\partial y_s^2}$ denotes the Laplacian operator.

The strong coupling dynamics and implicit behavior in~\eqref{s5.3.2eq1sub1} make the data-driven discovery of the simplified BZ reaction challenging when using implicit-SINDy and PDE-FIND. However, SINDy-PI correctly identifies the simplified dynamics of the BZ-Reaction, as shown in Fig.~\ref{BZ_Reaction}.  
To generate the simplified BZ reaction data, we use a spectral method~\cite{Trefethen_Book,kutz2013data} with time horizon $T=1$ and time step of $dt=0.001$.   
We use $n=128$ discretization points with spatial domain ranging from $-10$ to $10$. The initial condition is chosen to be a mixture of Gaussian functions. 
$80\%$ of the data is used for training, and the remaining $20\%$ is used for model selection. 
The right-hand side library is normalized during the sparse regression process. A range of sparsity parameters $\lambda$ are tested from $0.1$ to $1$, with increments of $0.1$ 
The other system parameters in~\eqref{s5.3.2eq1} are given in App.~\ref{E1} and the SINDy-PI model is given in App.~\ref{App:SINDyBZ}. 


\subsection{Extracting Physical Laws and Conserved Quantities}
\label{sec6}
In this final example, we demonstrate how to use SINDy-PI to extract governing physical laws and conserved quantities from data. 
Many systems of interest are governed by Hamiltonian or Lagrangian dynamics. 
Instead of identifying the ODE or PDE equations of motion, it might be possible to extract the physical laws directly. 
These equations contain important information about the system and may be more concise, useful, and straightforward than the underlying ODE or PDE. 
For example, given a Lagrangian, we can  derive the equations of motion. 

The most difficult aspect of using SINDy-PI to identify a physical law is how to build the library. Conservation laws may contain higher-order derivatives, such as $\ddot{x}$. To include all possible terms, the library may become exceedingly large.  
The library size will also increase if the system has many states. 
Large libraries make the sparse regression sensitive to noise. Thus, extracting the physical law from data using SINDy-PI is still challenging due to the lack of constraints when constructing the library function. We only show one example in our paper to demonstrate that it is possible to achieve this using SINDy-PI, but further work is required to reduce the library size so that the sparse regression is robust.

\begin{figure}[t]
    \vspace{-.2in}
    \centering
    \includegraphics[width=1\textwidth]{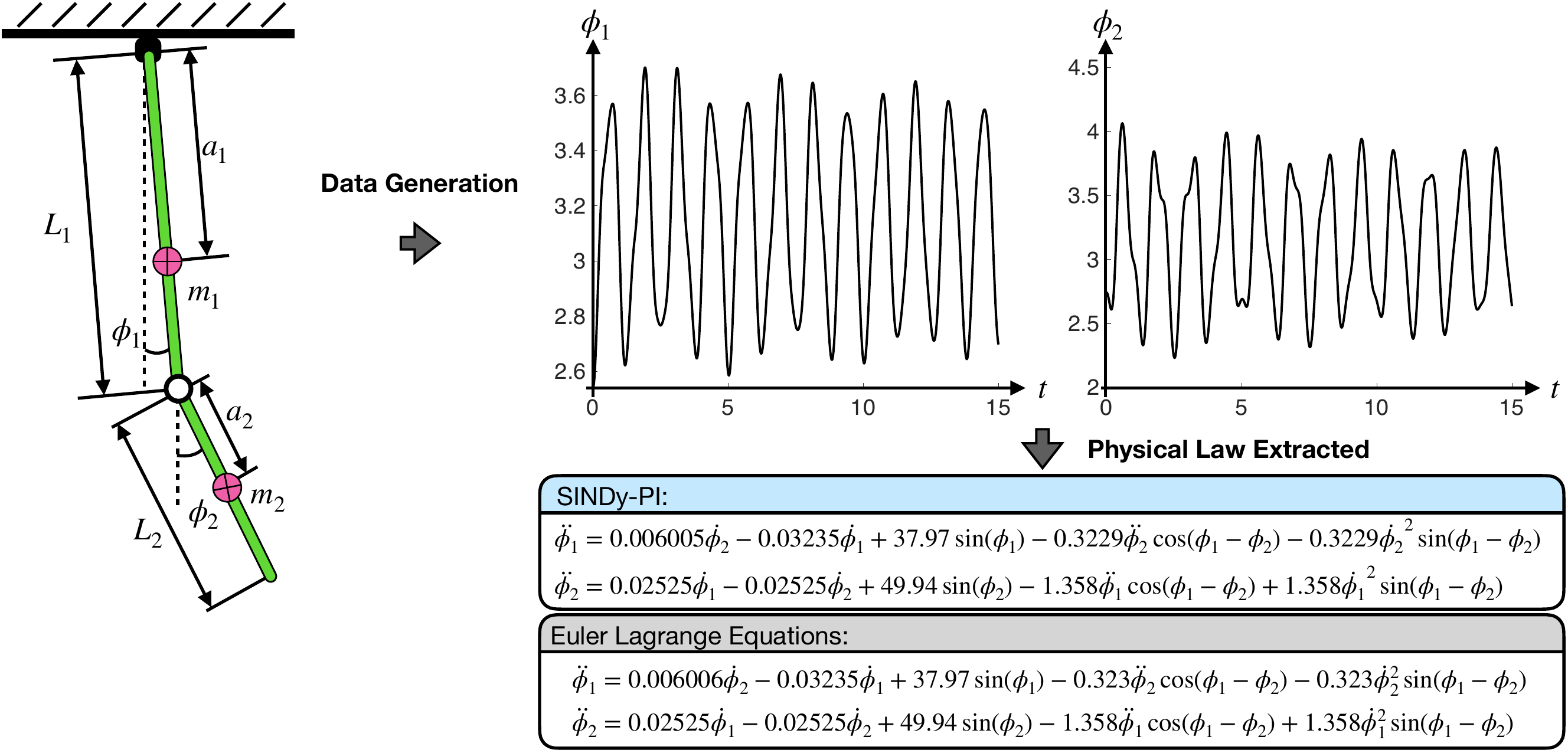}
    \caption{SINDy-PI is used to extract the conserved quantity for a double pendulum.}
    \label{ExtractLaw}
\end{figure}

As an example, we consider the double pendulum shown in Fig.~\ref{ExtractLaw}, with the system parameters given in App.~\ref{E1}. In this case, we also account for the friction in the pendulum joint, with friction constants of $k_1=7.2484\times 10^{-4}$ and $k_2=1.6522\times 10^{-4}$ for the pendulum arms, respectively. 
In this case, we extract the Lagrangian of the double pendulum~\cite{Graichen2007} using SINDy-PI. 
To extract this Lagrangian, we simulate the system with initial condition $x_{\text{train}}=[\pi-0.6,\ \pi-0.4,\ 0,\ 0]^T$ from $t=0$ to $t=15$  with time step $dt=0.001$. The resulting model is shown in Fig.~\ref{ExtractLaw}.

\section{Conclusions and Future Work}
\label{sec7}
In this paper, we develop SINDy-PI (parallel,implicit), a robust variant of the SINDy algorithm to identify implicit dynamics and rational nonlinearities. 
SINDy-PI overcomes the sensitivity of the previous implicit-SINDy approach, which is based on a null-space calculation, making it highly sensitive to noise.  
Instead, we introduce both parallel and constrained optimizations to test candidate terms in the dynamics, making the new SINDy-PI algorithm as robust as the original SINDy algorithm.  
We also extend the algorithm to incorporate external forcing and actuation, making it more applicable to real-world systems.  
We demonstrate this approach on several challenging systems with implicit and rational dynamics, including ODEs, actuated systems, and PDEs.  
In particular, we discover the implicit dynamics for a simplified model for the BZ chemical reaction PDE, the double pendulum mechanical system, and the yeast glycolisis model, which have all been challenging test cases for advanced identification techniques.  
Throughout these examples, we demonstrate considerable noise robustness and reductions to the data required, over the previous implicit-SINDy algorithm.

Despite the advances outlined here, there are still many important avenues of future work. 
One limitation of this approach, and of SINDy in general, is in the design of the library of candidate functions.  The goal is a descriptive library, but the library size grows rapidly, which in turn makes the sparse regression ill-conditioned; other issues effecting robustness are discussed in App.~\ref{App:Robustness}. Recently, tensor approaches have been introduced to alleviate this issue, making libraries both descriptive and tractable~\cite{Gelss2019mindy}, and this is a promising approach that may be incorporated in SINDy-PI as well.  
More generally, automatic library generation, guided by expert knowledge, is an important topic.  
Other research directions will involve parameterizing elements of the library, so that the algorithm simultaneously identifies the model structure and the parameters of the sparsely selected terms.  
Recent unified optimization frameworks, such as SR3~\cite{ZhengSR3,Champion2019SINDySR3}, may make this possible.  
Model selection is another key area that will required focused attention.  
Balancing accuracy on test data, sparsity of the model, and the potential for overfitting are all serious concerns.  
The sparse regression and optimization may also be improved for better noise robustness.  
Finally, modifying SINDy-PI to incorporate prior physical knowledge and to only model the discrepancy with an existing model~\cite{KK2019Discrepancy} will be the focus of ongoing work.  

\section*{Acknowledgments}
SLB acknowledges support from the Army Research Office (ARO W911NF-19-1-0045) and the Air Force Office of Scientific Research (AFOSR FA9550-18-1-0200). 
JNK acknowledges support from the Air Force Office of Scientific Research (AFOSR FA9550-17-1-0329). 
We also acknowledge valuable discussions with Aditya Nair, Eurika Kaiser, Brian DeSilva, Tony Piaskowy, Jared Callaham, and Benjamin Herrmann. We thank Ariana Mendible for reviewing the manuscript and providing useful suggestions.


 \begin{spacing}{.8}
 \small{
 \setlength{\bibsep}{1.5pt}
\bibliographystyle{IEEEtran}
\bibliography{PaperReference}
 }
 \end{spacing}


\appendix
\section{Noise Sensitivity of SINDy-PI and implicit-SINDy}
\label{A1}
\subsection{Performance Evaluation Criteria}
\label{Appen_PEC}
To compare the performance of SINDy-PI and implicit-SINDy for noisy data, we must define an evaluation criteria. 
We compare the performance of the best model generated by each method that has the lowest prediction error on the test data, selected according to Eq.~\eqref{eq_n1}. 
To compare the models generated by the two methods with the ground truth model, we use the concept of model discrepancy~\cite{KK2019Discrepancy,M.C.Kennedy2001BayesianCalibration,Arendt2012ModelUncertainty} and set prediction accuracy, structural accuracy, and parameter accuracy as our performance criteria. A good prediction error does not guarantee the model has good structural accuracy and parameter accuracy, and vice versa, motivating multiple performance criteria.


\subsection{Numerical Experiments}\label{App:NumericalExperiments}


We use the Michaelis–Menten kinetics,  given by Eq.~\eqref{s4eq1}, to compare the performance of SINDy-PI and implicit-SINDy. We performed our numerical experiments as follows:
\begin{enumerate}[Step 1:]
    \item Randomly generate 2400 different initial conditions of different magnitudes ranging from $0$ to $12.5$. Simulate those initial conditions using a fourth-order Runge-Kutta method with time step $dt=0.1$ and time horizon $T=5$. The testing data is generated using $600$ random initial conditions using the same method as the training data.
    \item Add Gaussian noise to the training and testing data. $23$ different Gaussian noise levels with magnitudes ranging from $10^{-7}$ to $5\times10^{-1}$ are used. For each noise level, $30$ different random noise realizations are generated, resulting in $30$ different noisy data sets for each noise level.
    \item Compute the derivative of the noisy data.  We investigate several approaches, including finite-difference and total-variation regularized difference (TVRegDiff)~\cite{Chartrand2011TVReg} derivatives.  In all cases, SINDy-PI is several orders of magnitude more robust to noise than implicit-SINDy, and only the result of using TVRegDiff is shown in this paper.  
    TVRegDiff generates more accurate derivatives, but also requires hyperparameter tuning and causes aliasing, so we trim the ends of the time series generated by each initial condition (first and last $30\%$). 
    It is possible to add Gaussian noise to the clean derivative data to investigate robustness, although this is less relevant for real-world scenarios, where only noisy state data is available.

    \item Train SINDy-PI and implicit-SINDy models on noisy training data. For each noise level, we sweep through $68$ different sparsity parameters $\lambda$ for SINDy-PI, from $0.01$ to $5$.  The $\lambda$ is varied by a factor of 2~\cite{Qu2014ADM} to calculate the null space in the implicit-SINDy method. The library for implicit-SINDy and SINDy-PI is
    \begin{equation}
        \label{s4eq2}
        \bTheta(\bX,\dot{\bX})=[\mathbf{1}\ \bX\ \bX^2\ \bX^3\ \bX^4\ \dot{\bX}\ \dot{\bX}\bX\ \dot{\bX}\bX^2\ \dot{\bX}\bX^3\ \dot{\bX}\bX^4].
    \end{equation}
    
    \item Due to the various parameter values, we use model selection to choose a model.  We use the test data with the same noise magnitude to perform the model selection process. The ratio of training data and testing data is $8:2$. 
    
    \item The best model generated by the two methods are compared. We use the prediction error, error in the model structure (i.e., the number of terms that are incorrectly present or missing from the model), and parameter error as our model performance evaluation criteria. We average the performance over $30$ different noise realizations for each noise level. We then plot the distribution of structure error in Fig.~\ref{TVReg_k0}.
    
\end{enumerate}


Many parameters affect the performance of these methods: the length of training data, prediction steps to calculate prediction error, the initial conditions for training data, choice of the library, and the derivative computation. 
We have attempted to carefully optimize each method, although an exhaustive parameter sweep is beyond the scope of the present work.  However, in all cases SINDy-PI outperforms implicit-SINDy.

\section{Data Usage of SINDy-PI and implicit-SINDy}
\label{B1}

Sec.~\ref{sec3.4} investigates the data usage of SINDy-PI and implicit-SINDy on the yeast glycolysis model in Eq.~\eqref{s4eq4}.  The parameters of this problems are given in Table.~\ref{Yeast_Glycolysis_Prameters}.
The data usage comparison is performed by the following steps:
\begin{enumerate}[Step 1:]
    \item Generate training data by simulating Eq.~\eqref{s4eq4} with parameters in Table.~\ref{Yeast_Glycolysis_Prameters} and  a time step of $dt=0.1$, with time horizon $T=5$.
    We simulate the system using $900$ random initial conditions with magnitude ranging from $0$ to $3$. 
    \item Shuffle the training data and  select $j$ percent of the  entire training data set at random to train the  SINDy-PI and implicit-SINDy models. These training data are sampled from all trajectories, and they are not necessarily consecutive in time. 
    No noise is added  since we only care about the effect of the data length in this case. The sparsity parameter $\lambda$ is fixed for both algorithms (different values); this value is selected for a single percentage $j$ where both methods fail to identify the correct model, and we sweep through $\lambda$.
    \item Run the numerical experiment for $20$ times for each data length and calculate the percentage of times the two algorithms yield the correct structure of the Eq.~\eqref{s4eq4sub6}.
\end{enumerate}

\begin{table}[]
\caption{The parameter used for simulating the Eq.~\eqref{s4eq4}.}
\vspace{-0.1in}
\label{Yeast_Glycolysis_Prameters}
\resizebox{\textwidth}{!}{%
\begin{tabular}{|c|c|c|c|c|c|c|c|c|c|c|c|c|c|}
\hline
\textbf{Parameter} & $c_1$ & $c_2$ & $c_3$   & $d_1$ & $d_2$   & $d_3$ & $d_4$   & $e_1$ & $e_2$ & $e_3$ & $e_4$ & $f_1$ & $f_2$ \\ \hline
Value              & 2.5   & -100  & 13.6769 & 200   & 13.6769 & -6    & -6      & 6     & -64   & 6     & 16    & 64    & -13   \\ \hline
\textbf{Parameter} & $f_3$ & $f_4$ & $f_5$   & $g_1$ & $g_2$   & $h_1$ & $h_2$   & $h_3$ & $h_4$ & $h_5$ & $j_1$ & $j_2$ & $j_3$ \\ \hline
Value              & 13    & -16   & -100    & 1.3   & -3.1    & -200  & 13.6769 & 128   & -1.28 & -32   & 6     & -18   & -100  \\ \hline
\end{tabular}
}
\end{table}

The final comparison is shown in Fig.~\ref{Data_Length_Compare}. 
Data usage requirements for other state equations are given in Table~~\ref{Data_Length_Compare_2}; Fig.~\ref{Data_Length_Compare} shows results for the hardest equation to identify.  
The other equations require less data.  Normalizing the SINDy-PI library improves data learning rates as well.

\begin{table}[]
\centering
\caption{Comparison of data usage of SINDy-PI and implicit-SINDy on other states.}
\vspace{-0.1in}
\label{Data_Length_Compare_2}
\renewcommand{\arraystretch}{1}
\resizebox{\textwidth}{!}{%
\begin{tabular}{|cccccccccccc|}
\cline{2-12}
\multicolumn{1}{c|}{} &
\multicolumn{1}{||c|}{Equation} & \multicolumn{2}{c|}{Eq.~\eqref{s4eq4sub1}} & \multicolumn{2}{c|}{Eq.~\eqref{s4eq4sub2}} & \multicolumn{1}{c|}{Eq.~\eqref{s4eq4sub3}} & \multicolumn{1}{c|}{Eq.~\eqref{s4eq4sub4}} & \multicolumn{1}{c|}{Eq.~\eqref{s4eq4sub5}} & \multicolumn{2}{c|}{Eq.~\eqref{s4eq4sub6}} & Eq.~\eqref{s4eq4sub7} \\ \cline{2-12}

\multicolumn{1}{c|}{} &
\multicolumn{1}{||c|}{Library Order} & \multicolumn{2}{c|}{$6$} & \multicolumn{2}{c|}{$6$} & \multicolumn{1}{c|}{$3$} & \multicolumn{1}{c|}{$3$} & \multicolumn{1}{c|}{$3$} & \multicolumn{2}{c|}{$6$} & $3$ \\

\hline \hline 

\multirow{3}{*}{\shortstack{SINDy-PI\\un-normalized}} &
\multicolumn{1}{||c|}{Left-Hand Side} & \multicolumn{1}{c|}{$\dot{x}_1$} & \multicolumn{1}{c|}{$\dot{x}_1x^4_6$} & \multicolumn{1}{c|}{$\dot{x}_2$} & \multicolumn{1}{c|}{$\dot{x}_2x^4_6$} & \multicolumn{1}{c|}{$\dot{x}_3$} & \multicolumn{1}{c|}{$\dot{x}_4$} & \multicolumn{1}{c|}{$\dot{x}_5$} & \multicolumn{2}{c|}{$\dot{x}_6x^6_4$} & $\dot{x}_7$ \\ \cline{2-12}

\multicolumn{1}{|c|}{} &
\multicolumn{1}{||c|}{Threshold} & 
\multicolumn{1}{c|}{$0.5$} & 
\multicolumn{1}{c|}{$0.05$} & 
\multicolumn{1}{c|}{$0.5$} & 
\multicolumn{1}{c|}{$0.05$} & 
\multicolumn{1}{c|}{$0.2$} & 
\multicolumn{1}{c|}{$0.5$} & 
\multicolumn{1}{c|}{$0.3$} & 
\multicolumn{2}{c|}{$0.01$} & $0.5$ \\ \cline{2-12}

\multicolumn{1}{|c|}{} &
\multicolumn{1}{||c|}{Data Usage} & \multicolumn{1}{c|}{$50\%$} & \multicolumn{1}{c|}{$7.5\%$} & \multicolumn{1}{c|}{$55\%$} & \multicolumn{1}{c|}{$8.5\%$} & \multicolumn{1}{c|}{$0.5\%$} & \multicolumn{1}{c|}{$0.5\%$} & \multicolumn{1}{c|}{$0.3\%$} & \multicolumn{2}{c|}{$40\%$} & $0.5\%$ \\ 

\hline \hline

\multirow{3}{*}{\shortstack{SINDy-PI\\normalized}} &
\multicolumn{1}{||c|}{Left-Hand Side} & \multicolumn{1}{c|}{$\dot{x}_1$} & \multicolumn{1}{c|}{$\dot{x}_1x^4_6$} & \multicolumn{1}{c|}{$\dot{x}_2$} & \multicolumn{1}{c|}{$\dot{x}_2x^4_6$} & \multicolumn{1}{c|}{$\dot{x}_3$} & \multicolumn{1}{c|}{$\dot{x}_4$} & \multicolumn{1}{c|}{$\dot{x}_5$} & \multicolumn{1}{c|}{$\dot{x}_6$} & \multicolumn{1}{c|}{$\dot{x}_6x^6_4$} & $\dot{x}_7$ \\ \cline{2-12}

\multicolumn{1}{|c|}{} &
\multicolumn{1}{||c|}{Threshold} & \multicolumn{1}{c|}{$0.5$} & \multicolumn{1}{c|}{$0.5$} & \multicolumn{1}{c|}{$0.5$} & \multicolumn{1}{c|}{$0.5$} & \multicolumn{1}{c|}{$0.6$} & \multicolumn{1}{c|}{$0.8$} & \multicolumn{1}{c|}{$0.4$} & \multicolumn{1}{c|}{$0.1$} & \multicolumn{1}{c|}{$0.1$} & $0.2$ \\ \cline{2-12}

\multicolumn{1}{|c|}{} &
\multicolumn{1}{||c|}{Data Usage} & \multicolumn{1}{c|}{$18\%$} & \multicolumn{1}{c|}{$3\%$} & \multicolumn{1}{c|}{$10\%$} & \multicolumn{1}{c|}{$3\%$} & \multicolumn{1}{c|}{$0.45\%$} & \multicolumn{1}{c|}{$0.45\%$} & \multicolumn{1}{c|}{$0.275\%$} & \multicolumn{1}{c|}{$35\%$} & \multicolumn{1}{c|}{$8\%$} & $0.4\%$ \\ 

\hline \hline


\multirow{2}{*}{\shortstack{implicit-SINDy\\normalized}} &
\multicolumn{1}{||c|}{Threshold} & \multicolumn{2}{c|}{$5\times 10^{-3}$} & \multicolumn{2}{c|}{$2\times 10^{-3}$} & \multicolumn{1}{c|}{$8\times 10^{-3}$} & \multicolumn{1}{c|}{$8\times 10^{-3}$} & \multicolumn{1}{c|}{$8\times 10^{-3}$} & \multicolumn{2}{c|}{$3\times 10^{-3}$} & $8\times 10^{-3}$ \\ \cline{2-12}

\multicolumn{1}{|c|}{} &
\multicolumn{1}{||c|}{Data Usage} & \multicolumn{2}{c|}{$10\%$} & \multicolumn{2}{c|}{$10\%$} & \multicolumn{1}{c|}{$0.5\%$} & \multicolumn{1}{c|}{$0.6\%$} & \multicolumn{1}{c|}{$0.3\%$} & \multicolumn{2}{c|}{$100\%$} & $0.5\%$ \\ \hline

\end{tabular}%
}
\end{table}

\section{SINDy-PI and PDE-FIND on Rational PDE Problem}
\label{D1}

 In Sec.~\ref{sec3.5}, we compared the performance of SINDy-PI and PDE-FIND on a modified KdV equation. 
 The simulation data is obtained using a spectral method~\cite{Trefethen_Book} with a time step  of $dt=0.01$ and time horizon $T=20$, spatial domain $L=-25\ \text{to}\ 25$, and $n=128$ spatial discretization points. The library of PDE-FIND is chosen to be
\begin{equation}
\begin{aligned}
    \label{s5.3.1eq3}
    \bTheta({\mathbf{U},\mathbf{U}_{x},\mathbf{U}_{xx},\mathbf{U}_{xxx}})=&[\mathbf{1}\ \mathbf{U}\ \mathbf{U}_x\ \mathbf{U}_{xx}\ \mathbf{U}_{xxx}\ \mathbf{U}_x^2\ \mathbf{U}_{xx}^2\ \mathbf{U}_{xxx}^2\ \mathbf{U}\mathbf{U}_x\ 
    \\ 
    &\mathbf{U}\mathbf{U}_{xx}\ \mathbf{U}\mathbf{U}_{xxx}\ \mathbf{U}^2\mathbf{U}_x^2\ \mathbf{U}^2\mathbf{U}_{xx}^2\ \mathbf{U}^2\mathbf{U}_{xxx}^2]
 \end{aligned}   
\end{equation}
and the right-hand side library for the SINDy-PI is chosen to be 
\begin{equation}
\begin{aligned}
    \label{s5.3.1eq4}
    \bTheta({\mathbf{U},\mathbf{U}_{x},\mathbf{U}_{xx},\mathbf{U}_{xxx}})=&[\mathbf{1}\ \mathbf{U}\ \mathbf{U}_t\ \mathbf{U}_x\ \mathbf{U}_{xx}\ \mathbf{U}_{xxx}\ \mathbf{U}^2\ \mathbf{U}_t^2\ \mathbf{U}_x^2\ \mathbf{U}_{xx}^2\ \mathbf{U}_{xxx}^2
    \\ 
    &\mathbf{U}\mathbf{U}_t\
    \mathbf{U}\mathbf{U}_x\ \mathbf{U}\mathbf{U}_{xx}\ \mathbf{U}\mathbf{U}_{xxx}\ \mathbf{U}^2\mathbf{U}_x^2\ \mathbf{U}^2\mathbf{U}_{xx}^2\ \mathbf{U}^2\mathbf{U}_{xxx}^2],
 \end{aligned}   
\end{equation}
while the left-hand side library is chosen to be
\begin{equation}
    \label{s5.3.1eq4}
    C(\mathbf{U},\mathbf{U}_t,\mathbf{U}_{x},\mathbf{U}_{xx})=[\mathbf{U}_t\ \mathbf{U}\mathbf{U}_{t}\ \mathbf{U}\mathbf{U}_{x}\ \mathbf{U}\mathbf{U}_{xx}].
\end{equation}

For both SINDy-PI and PDE-FIND, we used $100$ different values for the sparsity parameter $\lambda$ ranging from $0.1$ to $10$ with step size $0.1$. We use $80\%$ of the simulation data for  training and $20\%$ for testing and model selection. We calculate the normalized prediction error for all models on state $u_t$ and the model with minimum prediction error is selected as the final model.

\section{Parameter Values for Simulations}
\label{E1}

The parameters for the double pendulum simulation in Sec.~\ref{secDoublePen} are given in Table.~\ref{DoublePenParameterTable}. 
The parameters used to simulate the simplified model of the Belousov-Zhabotinsky reaction in Eq.~\eqref{s5.3.2eq1} are given in Table.~\ref{BZ_Parameters}.

\begin{table}[h]
\centering
\caption{Parameters used to simulate the double pendulum.} 
\vspace{-0.1in}
\label{DoublePenParameterTable}
\begin{tabular}{|c|c|c|c|c|c|c|c|c|c|}
\hline
\textbf{Parameter} & $m_1$ & $m_2$ & $L_1$ & $L_2$ & $a_1$ & $a_2$ & $I_1$ & $I_2$ & $g$ \\ \hline
Value & $0.2704$ & $0.2056$ & $0.2667$ & $0.2667$ & $0.191$ & $0.1621$ & $0.003$ & $0.0011$ & $9.81$ \\ \hline
\end{tabular}%
\end{table}

\begin{table}[h]
\centering
\caption{Parameters used to simulate the single pendulum on a cart.} 
\vspace{-0.1in}
\label{SinglePenParameterTable}
\begin{tabular}{|c|c|c|c|c|}
\hline
\textbf{Parameter} & $m$ & $L$ & $M$ & $g$  \\ \hline
Value & $1$ & $1$ & $1$ &$9.81$ \\ \hline
\end{tabular}%
\end{table}

\begin{table}[h]
\centering
\caption{Parameters Used in Eq.~\eqref{s5.3.2eq1} for Simulating the Belousov-Zhabotinsky Reaction Model.}
\vspace{-0.1in}
\label{BZ_Parameters}
\begin{tabular}{|cccccccccccccc|}
\hline
\multicolumn{1}{|c|}{\textbf{Parameter}} & \multicolumn{1}{c|}{$q$} & \multicolumn{1}{c|}{$f$} & \multicolumn{1}{c|}{$\varepsilon$} & \multicolumn{1}{c|}{$\alpha$} & \multicolumn{1}{c|}{$\beta$} & \multicolumn{1}{c|}{$\gamma$} & \multicolumn{1}{c|}{$\varepsilon_2$} & \multicolumn{1}{c|}{$\varepsilon_3$} & \multicolumn{1}{c|}{$\chi$} & \multicolumn{1}{c|}{$D_x$} & \multicolumn{1}{c|}{$D_z$} & \multicolumn{1}{c|}{$D_s$} & $D_u$ \\ \hline
\multicolumn{1}{|c|}{Value} & \multicolumn{1}{c|}{$1$} & \multicolumn{1}{c|}{$1.5$} & \multicolumn{1}{c|}{$0.3$} & \multicolumn{1}{c|}{$0.3$} & \multicolumn{1}{c|}{$0.26$} & \multicolumn{1}{c|}{$0.4$} & \multicolumn{1}{c|}{$0.15$} & \multicolumn{1}{c|}{$0.03$} & \multicolumn{1}{c|}{$0$} & \multicolumn{1}{c|}{$0.01$} & \multicolumn{1}{c|}{$0.01$} & \multicolumn{1}{c|}{$1$} & $1$ \\ \hline
\end{tabular}%
\end{table}
\section{SINDy-PI Models for the Single Pendulum on a Cart}\label{App:SINDyPendulumCart}
The Lagrangian for the single pendulum on a cart with an input force on the cart is:
\begin{equation}
    \mathcal{L}=T-V=\frac{1}{2}(m+M) \dot{s}^2 + \frac{1}{2}L^2m \dot{\phi}^2 - Lgm \cos(\phi) + Lm \cos(\phi) \dot{\phi} \dot{s},
\end{equation}
where $m$ is the mass at the end of the pendulum arm, $M$ is the mass of the cart, $L$ is the length of the pendulum arm, $s$ is the position of the cart, and $\phi$ is the pendulum angle. 
We do not consider damping in this case. 
Using the numeric values $m=M=L=1$ and $g=-9.81$ this simplifies to
\begin{equation}
    \label{SingleLag}
    \mathcal{L}=T-V=\dot{s}^2 + \frac{1}{2} \dot{\phi}^2 - 9.81 \cos(\phi) + \cos(\phi) \dot{\phi} \dot{s},
\end{equation}
The Euler-Lagrange equation of the system are
\begin{subequations}
    \label{SingleEL}
    \begin{align}
  \begin{split}  \frac{d}{dt}\frac{\partial{\mathcal{L}}}{\partial {\dot{\phi}}}-\frac{\partial{\mathcal{L}}}{\partial{\phi}}&=0,\\
    \frac{d}{dt}\frac{\partial{\mathcal{L}}}{\partial {\dot{s}}}-\frac{\partial{\mathcal{L}}}{\partial{s}}&=F,
    \end{split} \quad\Longrightarrow\quad
    \begin{split}
            m L^2\ddot{\phi} + m L\ddot{s}\cos(\phi) - L g m \sin(\phi) &= 0\\
    (M+m)\ddot{s} - F - mL\sin(\phi)\dot{\phi}^2 + mL\ddot{\phi}\cos(\phi) &= 0
    \end{split}
    \end{align}
\end{subequations}
where $F$ is the force applied to the pendulum cart.
It is possible to isolate $\ddot{\phi}$ and $\ddot{s}$:
\begin{subequations}
    \begin{align}
        \ddot{\phi}&=\frac{-(F\cos(\phi) - Mg\sin(\phi) - mg\sin(\phi) + Lm\cos(\phi)\sin(\phi)\dot{\phi}^2)}{L(M + m\sin(\phi)^2)}\label{SingleEL_a_a_a},
        \\
        \ddot{s}&=\frac{F + Lm\sin(\phi)\dot{\phi}^2 - mg\cos(\phi)\sin(\phi)}{M + m\sin(\phi)^2}\label{SingleEL_b_b_b}.
    \end{align}
\end{subequations}

It is possible to write this as a system of four coupled first-order equations
\begin{subequations}
    \begin{align}
        \frac{d}{dt}\phi&=\dot{\phi},\\
        \frac{d}{dt}s&=\dot{s},\\
        \frac{d}{dt}\dot{\phi}&=\frac{-(F\cos(\phi) - Mg\sin(\phi) - mg\sin(\phi) + Lm\cos(\phi)\sin(\phi)\dot{\phi}^2)}{L(M + m\sin(\phi)^2)},\\
        \frac{d}{dt}\dot{s}&=\frac{F + Lm\sin(\phi)\dot{\phi}^2 - mg\cos(\phi)\sin(\phi)}{M + m\sin(\phi)^2}.
    \end{align}
\end{subequations}
With the numerical values shown in Table.~\ref{SinglePenParameterTable}, this becomes 
\begin{subequations}
    \label{Single_Pendulum_Numerical}
    \begin{align}
        \frac{d}{dt}\phi&=\dot{\phi},\\
        \frac{d}{dt}s&=\dot{s},\\
        \frac{d}{dt}\dot{\phi}&=\frac{19.62\sin{(\phi)}-F\cos{(\phi)}-\sin{(\phi)}\cos{(\phi)}\dot{\phi}^2}{2-\cos{(\phi)}^2},
        \label{SinglePenSubEq3}\\
        \frac{d}{dt}\dot{s}&=\frac{2F-9.81\sin{(2\phi)}+2\sin{(\phi)}\dot{\phi}^2}{2+2\sin{(\phi)}^2}\label{SinglePenSubEq4}.
    \end{align}
\end{subequations}
Parameters identified by SINDy-PI under different noise magnitudes are presented in Tables~\ref{SingleParameterSubEq3} and~\ref{SingleParameterSubEq4}.

\begin{table}[h]
\centering
\caption{Parameters identified by SINDy-PI for Eq.~\eqref{SinglePenSubEq3} under different noise magnitudes.}
\vspace{-0.1in}
\label{SingleParameterSubEq3}
\begin{tabular}{|c||c|c|c|c|c|}
\hline
\multirow{4}{*}{\diagbox[height=4\line,width=10em]{Noise\\Magnitude}{Value}{Basis}} & \multicolumn{3}{c|}{\multirow{2}{*}{Numerator}}                                                                              & \multicolumn{2}{c|}{\multirow{2}{*}{Denominator}}             \\
                       & \multicolumn{3}{c|}{}                                                                                                        & \multicolumn{2}{c|}{}                                         \\ \cline{2-6} 
                       & \multirow{2}{*}{$\sin{(\phi)}$} & \multirow{2}{*}{$F\cos{(\phi)}$} & \multirow{2}{*}{$\sin{(\phi)}\cos{(\phi)}\dot{\phi}^2$} & \multirow{2}{*}{Constant} & \multirow{2}{*}{$\cos{(\phi)}^2$} \\
                       &                                 &                                  &                                                         &                           &                                   \\ \hline \hline
0                      & 19.62                           & -1                               & -1                                                      & 2                         & -1                                \\ \hline
0.001                  & 19.618                          & -1.0005                          & -0.9999                                                 & 2                         & -1                                \\ \hline
0.005                  & 19.6135                         & -1.171                           & -0.9996                                                 & 2                         & -0.9997                           \\ \hline
0.02                   & 19.5881                         & Not Identified                   & -0.4912                                                 & 2                         & -1.0122                           \\ \hline
\end{tabular}
\end{table}

\begin{table}[h]
\centering
\caption{Parameters identified by SINDy-PI for Eq.~\eqref{SinglePenSubEq4} under different noise magnitudes.}
\vspace{-0.1in}
\label{SingleParameterSubEq4}
\begin{tabular}{|c||c|c|c|c|c|}
\hline
\multirow{4}{*}{\diagbox[height=4\line,width=10em]{Noise\\Magnitude}{Value}{Basis}} & \multicolumn{3}{c|}{\multirow{2}{*}{Numerator}}                                                                              & \multicolumn{2}{c|}{\multirow{2}{*}{Denominator}}             \\
                       & \multicolumn{3}{c|}{}                                                                                                        & \multicolumn{2}{c|}{}                                         \\ \cline{2-6} 
                       & \multirow{2}{*}{$F$} & \multirow{2}{*}{$\sin{(2\phi)}$} & \multirow{2}{*}{$\sin{(\phi)}\dot{\phi}^2$} & \multirow{2}{*}{Constant} & \multirow{2}{*}{$\sin{(\phi)}^2$} \\
                       &                                 &                                  &                                                         &                           &                                   \\ \hline \hline
0                      & 2                           & -9.81                               & 2                                                      & 2                         & 2                                \\ \hline
0.001                  & 1.9992                          & -9.816                          & 1.9992                                                 & 2                         & 1.9992                                \\ \hline
0.005                  & 1.9982                         & -9.8015                           & 1.9986                                                 & 2                         & 1.9986                          \\ \hline
0.02                   & 2.0705                         & -9.8234                   & 2.0041                                                 & 2                         & 2.0041                          \\ \hline
\end{tabular}
\end{table}
\section{SINDy-PI Models for the Mounted Double Pendulum}\label{App:SINDyDoublePendulum}
For a mounted double pendulum system shown in Fig.~\ref{DoublePenFig} we could have following parameters: the parameters of the pendulum are center of mass $m_1$ and $m_2$, center of mass position $a_1$ and $a_2$, arm length $L_1$ and $L_2$, arm inertia $I_1$ and $I_2$, arm rotational angle $\phi_1$ and $\phi_2$, gravity acceleration $g$. Those values could be seen from Table.~\ref{DoublePenParameterTable}. If we consider friction between the pendulum joint, we could define $k_1=7.2485\times 10^{-4}$ and $k_2=1.6522\times 10^{-4}$ as our damping coefficient.It is easy to derive the Lagrangian of the mounted double pendulum which is given by
\begin{equation}
\begin{split}
        \mathcal{L}=T-V=&
        (m_2((L_1\cos(\phi_1)\dot{\phi}_1 +a_2\cos(\phi_2)\dot{\phi}_2)^2 +(L_1\sin(\phi_1)\dot{\phi}_1 + a_2\sin(\phi_2)\dot{\phi}_2)^2))/2 +
        \\
        &(m_1(a_1^2\cos(\phi_1)^2\dot{\phi}_1^2 + a_1^2\sin(\phi_1)^2\dot{\phi}_1^2))/2 + (I_1\dot{\phi}_1^2)/2 + (I_2\dot{\phi}_2^2)/2 
        - gm_2(a_2\cos(\phi_2)
        \\
        &+ L_1\cos(\phi_1)) - a_1gm_1\cos(\phi_1)
\end{split}
\end{equation}
The damping term caused by friction with friction coefficients $k_1$ and $k_2$ is 
\begin{equation}
    R_a=\frac{1}{2}k_1\dot{\phi}_1+\frac{1}{2}k_2(\dot{\phi}_1-\dot{\phi}_2)^2
\end{equation}
 The Euler-Lagrange equations with a Rayleigh dissipation term are then:
\begin{subequations}
    \begin{align}
    \frac{d}{dt}\frac{\partial{\mathcal{L}}}{\partial {\dot{\phi}_1}}-\frac{\partial{\mathcal{L}}}{\partial{\phi_1}}+\frac{\partial{R_a}}{\partial{\dot{\phi}_1}}&=0\label{DoubleEL_a},
    \\
    \frac{d}{dt}\frac{\partial{\mathcal{L}}}{\partial {\dot{\phi}_2}}-\frac{\partial{\mathcal{L}}}{\partial{\phi_2}}+\frac{\partial{R_a}}{\partial{\dot{\phi}_2}}&=0\label{DoubleEL_b}.
    \end{align}
\end{subequations}
The symbolic form of the Eq.~\eqref{DoubleEL_a} is
\begin{equation}
\begin{split}
\label{DoubleEL_a_a}
    &I_1\ddot{\phi}_1 + k_1\dot{\phi}_1 + k_2\dot{\phi}_1 +
    L_1^2\ddot{\phi}_1m_2 + a_1^2\ddot{\phi}_1m_1 + L_1a_2m_2\sin(\phi_1 - \phi_2)\dot{\phi}_2^2 
    \\
    &+L_1a_2\ddot{\phi}_2m_2\cos(\phi_1 - \phi_2) -k_2\dot{\phi}_2 - L_1gm_2\sin(\phi_1) -
    a_1gm_1\sin(\phi_1)=0,
\end{split}
\end{equation}
and the symbolic form of Eq.~\eqref{DoubleEL_b} is
\begin{equation}
\begin{split}
\label{DoubleEL_b_b}
    &I_2\ddot{\phi}_2 + k_2\dot{\phi}_2 + a_2^2\ddot{\phi}_2m_2 + L_1a_2\ddot{\phi}_1m_2\cos(\phi_1 - \phi_2) -k_2\dot{\phi}_1
    \\
    &- a_2gm_2\sin(\phi_2) - L_1a_2m_2\sin(\phi_1 - \phi_2)\dot{\phi}_1^2=0.
\end{split}
\end{equation}

Using the numerical parameter values in these equations gives 
\begin{align*}
    \ddot{\phi}_1 + 0.03235\dot{\phi}_1 + 0.323\ddot{\phi}_2\cos(\phi_1 - \phi_2) + 0.323\dot{\phi}_2^2\sin(\phi_1 - \phi_2) - 0.006006\dot{\phi}_2 - 37.97\sin(\phi_1) =0.\\
        \ddot{\phi}_2 + 0.02525\dot{\phi}_2 + 1.358\ddot{\phi}_1\cos(\phi_1 - \phi_2) - 0.02525\dot{\phi}_1 - 49.94\sin(\phi_2) - 1.358\dot{\phi}_1^2\sin(\phi_1 - \phi_2) =0.
\end{align*}
If we set $k_1=k_2=0$ and combine the equations, it is possible to solve for $\ddot{\phi}_1$ and $\ddot{\phi}_2$
\begin{equation*}
\begin{split}
\label{DoubleEL_EOM1}
    \ddot{\phi}_1&=
    (L_1a_2^2gm_2^2\sin(\phi_1) - 2L_1a_2^3\dot{\phi}_2^2m_2^2\sin(\phi_1 - \phi_2) + 
    2I_2L_1gm_2\sin(\phi_1) 
    \\
    &+ L_1a_2^2gm_2^2\sin(\phi_1 - 2\phi_2) + 2I_2a_1gm_1\sin(\phi_1) - L_1^2a_2^2\dot{\phi}_1^2m_2^2\sin(2\phi_1 - 2\phi_2) 
    \\
    &- 2I_2L_1a_2\dot{\phi}_2^2m_2\sin(\phi_1 - \phi_2) + 2a_1a_2^2gm_1m_2\sin(\phi_1))
    /(2I_1I_2 + L_1^2a_2^2m_2^2 
    \\
    &+ 2I_2L_1^2m_2 + 2I_2a_1^2m_1 + 2I_1a_2^2m_2 - L_1^2a_2^2m_2^2\cos(2\phi_1 - 2\phi_2) + 2a_1^2a_2^2m_1m_2)
\end{split}
\end{equation*}
and 
\begin{equation*}
\begin{split}
\label{DoubleEL_EOM2}
    \ddot{\phi}_2&=(a_2m_2(2I_1g\sin(\phi_2) + 2L_1^3\dot{\phi}_1^2m_2\sin(\phi_1 - \phi_2) + 2L_1^2gm_2\sin(\phi_2) 
    + 2I_1L_1\dot{\phi}_1^2\sin(\phi_1 - \phi_2) 
    \\
    &+ 2a_1^2gm_1\sin(\phi_2) + L_1^2a_2\dot{\phi}_2^2m_2\sin(2\phi_1 - 2\phi_2) 
    + 2L_1a_1^2\dot{\phi}_1^2m_1\sin(\phi_1 - \phi_2)
    \\
    &- 2L_1^2gm_2\cos(\phi_1 - \phi_2)\sin(\phi_1) - 2L_1a_1gm_1\cos(\phi_1 - \phi_2)\sin(\phi_1)))
    \\
    &/(2(I_1I_2 + L_1^2a_2^2m_2^2 + I_2L_1^2m_2 + I_2a_1^2m_1 + I_1a_2^2m_2 - L_1^2a_2^2m_2^2\cos(\phi_1 - \phi_2)^2 + a_1^2a_2^2m_1m_2)). 
\end{split}
\end{equation*}
If we use the values in Table.~\ref{DoublePenParameterTable} we have
\begin{equation*}
\begin{split}
    \label{DoublePenNum1}
    \ddot{\phi}_1&=(- 0.2808\sin(2\phi_1 - 2\phi_2)\dot{\phi}_1^2 - 0.4136\sin(\phi_1 - \phi_2)\dot{\phi}_2^2
    \\
    &+ 10.3278\sin(\phi_1 - 2\phi_2) + 38.2984\sin(\phi_1))/(1-0.2808\cos(2\phi_1 - 2\phi_2))
 ,
\end{split}
\end{equation*}
and
\begin{equation*}
\begin{split}
    \label{DoublePenNum2}
    \ddot{\phi}_2&=(1.7390\sin(\phi_1 - \phi_2)\dot{\phi}_1^2 + 0.2808\sin(2\phi_1 - 2\phi_2)\dot{\phi}_2^2 
    \\
    &- 33.02\sin(2\phi_1 - \phi_2) + 30.9472\sin(\phi_2))/(1-0.2808\cos(2\phi_1 - 2\phi_2)).
\end{split}
\end{equation*}

With no noise, SINDy-PI discovers the correct equations.  When we add random noise with magnitude of $0.005$, SINDy-PI discovers the following
\begin{equation*}
\begin{split}
    \label{DoublePenNum1_N1}
    \ddot{\phi}_1&=(- 0.2799\sin(2\phi_1 - 2\phi_2)\dot{\phi}_1^2 - 0.4137\sin(\phi_1 - \phi_2)\dot{\phi}_2^2 + 
    \\
    &+10.3429\sin(\phi_1 - 2\phi_2) + 38.3117\sin(\phi_1))/(1-0.2815\cos(2\phi_1 - 2\phi_2)),
\end{split}
\end{equation*}
and
\begin{equation*}
\begin{split}
    \label{DoublePenNum2_N1}
    \ddot{\phi}_2&=(1.7392\sin(\phi_1 - \phi_2)\dot{\phi}_1^2 + 0.2805\sin(2\phi_1 - 2\phi_2)\dot{\phi}_2^2\\
    & - 33.0035\sin(2\phi_1 - \phi_2)+ 30.9418\sin(\phi_2))/(1-0.2813\cos(2\phi_1 - 2\phi_2)).
\end{split}
\end{equation*}

If we increase the noise magnitude to $0.01$ then the SINDy-PI discovered equation becomes
\begin{equation*}
\begin{split}
    \label{DoublePenNum1_N1}
    \ddot{\phi}_1&=(- 0.2768\dot{\phi}_1^2\sin(2\phi_1 - 2\phi_2)- 0.4138\sin(\phi_1 - \phi_2)\dot{\phi}_2^2 \\
    & +10.3676\sin(\phi_1 - 2\phi_2) + 38.3225\sin(\phi_1))/(1-0.2818\cos(2\phi_1 - 2\phi_2)),
\end{split}
\end{equation*}
and
\begin{equation*}
\begin{split}
    \label{DoublePenNum2_N1}
    \ddot{\phi}_2&=(1.7355\sin(\phi_1 - \phi_2)\dot{\phi}_1^2 + 0.2794\sin(2\phi_1 - 2\phi_2)\dot{\phi}_2^2+0.1675\sin(2\phi_1 - 2\phi_2)  \\
    & - 33.0445\sin(2\phi_1 - \phi_2)+ 31.0065\sin(\phi_2))/(1-0.2819\cos(2\phi_1 - 2\phi_2)).
\end{split}
\end{equation*}
If we continue to increase the noise magnitude to $0.05$ then SINDy-PI incorrectly identifies
\begin{equation*}
\begin{split}
    \label{DoublePenNum1_N1}
    \ddot{\phi}_1&=(15.5413\sin(\phi_1 - 2\phi_2) - 2.6396\sin(\phi_1 - \phi_2) - 0.9538\cos(\phi_1 - \phi_2)
    \\
    &+ 35.9971\cos(\phi_1 - 61/40) 
    - 2.4160\cos(\phi_2 - 1149/1000) 
    - 0.0733\cos(2\phi_1 - 2\phi_2) 
    \\
    & + 0.0269\cos(4\phi_1 - 2\phi_2) +2.3419\sin(2\phi_1 - \phi_2) + 0.5142\dot{\phi}_1\sin(2\phi_1 - 2\phi_2) 
    \\
    &- 0.4584\dot{\phi}_2\sin(2\phi_1 - \phi_2) - 0.3807\dot{\phi}_2^2\sin(\phi_1 - \phi_2) + 
   0.8810\dot{\phi}_1\sin(\phi_1 - \phi_2)\\ &
   + 0.9411\dot{\phi}_1\sin(\phi_1 - 2\phi_2)
   - 0.7664\dot{\phi}_2\sin(\phi_1 - \phi_2) + 1.2026)/(1-0.4245\cos(2\phi_1 - 2\phi_2)),
\end{split}
\end{equation*}
and
\begin{equation*}
\begin{split}
    \label{DoublePenNum2_N1}
    \ddot{\phi}_2&=70.9\sin(\phi_1 - \phi_2).
\end{split}
\end{equation*}

\section{SINDy-PI Model for the Belousov-Zhabotinsky Reaction}\label{App:SINDyBZ}
The SINDy-PI discovered PDE for the simplified BZ reaction is
    \begin{align*}
        x_{\tau}&=\Delta{x}+\frac{0.24667x + 0.33333s + 0.5z  + 3.3333xs - 5.0xz + 2.1333x^2 - 3.3333x^3}{x + 0.1},
         \\
         z_{\tau}&=0.01\Delta{z}+x + 0.4u - 1.3z,
         \\
         s_{\tau}&=\Delta{s}+0.17333r - 0.66667s,
         \\
         u_{\tau}&=\Delta{u} - 133.33u + 100z.
    \end{align*}

\section{Inability to Identify Rational Dynamics with SINDy}
\begin{figure}[t]
    \vspace{-.1in}
    \centering
    \includegraphics[width=1\textwidth]{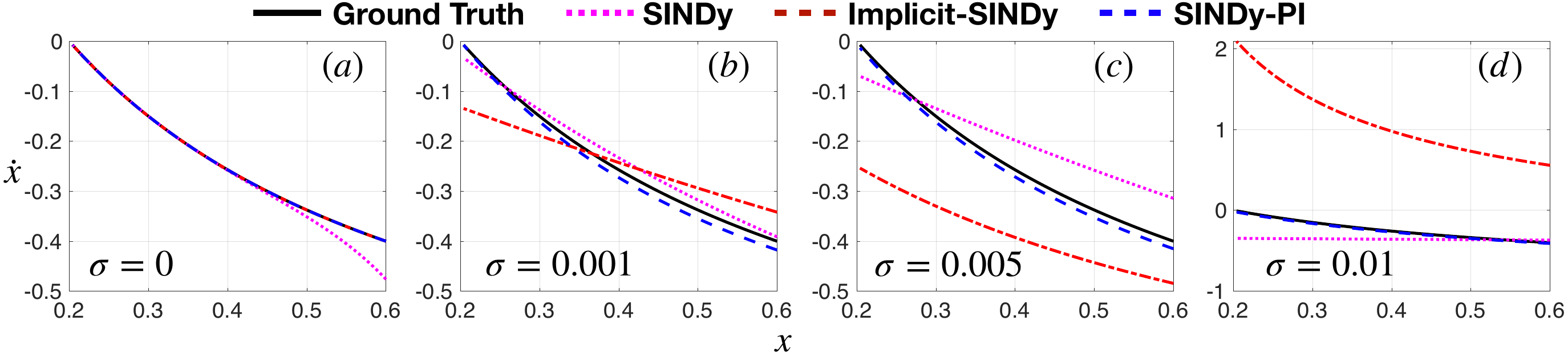}
    \caption{Comparison of the model identified by SINDy, Implicit-SINDy, and SINDy-PI on Mechaelis-Menten dynamics.}
    \label{TaylorResult}
\end{figure}

In this section, we demonstrate that it is not possible to identify rational dynamics with the original SINDy algorithm, testing it on the Michaelis-Menten dynamics in Eq.~\eqref{s4eq1}. We use the same parameters in Sec.~\ref{sec3.3}. 
The Taylor expansion of Eq.~\eqref{s4eq1} at $x=0$ is
\begin{equation}
\label{SINDyTaylorModel}
    \dot{x}\approx0.6-5x+\frac{50}{3}x^2-\frac{500}{9}x^3+\frac{5000}{27}x^4-\frac{50000}{81}x^5.
\end{equation}
Thus, when the trajectory provided for training is close enough to $x=0$, SINDy should identify Eq.~\eqref{SINDyTaylorModel}. 
To verify this, data with $x_0=0.2409$ is simulated for $22$ time steps with $dt=0.01$. Both the Implicit-SINDy and SINDy-PI algorithms identify the correct model in Eq.~\eqref{s4eq1} with highly accurate parameters. The model identified by SINDy is 
\begin{equation}
    \label{SINDyTaylorModelID}
    \dot{x}=0.5914-4.7387x+13.1389x^2-27.6470x^3+36.9846x^4-22.8388x^5.
\end{equation}
Thus, SINDy correctly identifies the first three terms of the Taylor expansion, although the higher order terms have large parameter errors.   
This model is compared with the SINDy-PI and Implicit-SINDy models in Fig.~\ref{TaylorResult} for a test trajectory initialized with $x_0=0.6$. 
From Fig.~\ref{TaylorResult} (a), it can be seen that both SINDy-PI and Implicit-SINDy match the true solution.  
However, the SINDy model only agrees for $x$ near the origin. When Gaussian noise of magnitude $\sigma$ is added, the SINDy model degrades further.  Moreover, the amount of data used needs to be increased to identify the correct model. In Fig.~\ref{TaylorResult} (b) to (d), $330$, $2200$ and $4400$ data points ranging from $0$ to $12$ are used for training. The same amount of data is used for model selection. From Fig.~\ref{TaylorResult} (b) to (d), it can be seen that the SINDy model does not work well away from $x=0$.

\section{Robustness of SINDy and SINDy-PI}\label{App:Robustness}
The robustness of SINDy-PI to noise depends on a number of factors, including the the length of the data, which initial conditions are chosen to generate the data, the model parameters, the order of the polynomial terms in the model, etc. The relative impact of all of these factors varies with the system we are studying. 

In general, training data that explores more of the phase space will result in a more robust model discovery process. 
Generally speaking, data that results in a better-conditioned  $\boldsymbol{\Theta}$ matrix will provide more robust results.  
Several studies have explored strategies to improve the robustness, with Wu and Xiu suggesting the use of a large ensemble of initial conditions~\cite{wu2018numerical}.  
Exciting transients is also important~\cite{Brunton2016SINDy}.  

Another key factor that affects the condition number of $\boldsymbol{\Theta}$ is the number of library elements, which is determined by the order for a polynomial library.  Including higher-order terms increases the condition number, making it more difficult to accurately disambiguate which nonlinear term is responsible for the observed behavior.  The library size scales exponentially with the maximum polynomial order.  

Noise robustness is also affected by the model parameters. 
Smaller parameters in $\boldsymbol{\Xi}$ are more likely to be removed during thresholding, making the procedure less robust to noise.  
To illustrate this, we change the parameter $K_m$ in Eq.~{s4eq1} and test the maximum noise SINDy-PI can handle. We set $K_m$ equal to $0.01$, $0.1$, $1$ and $10$ separately. We generate the training data with $120$ random initial conditions ranging from $0$ to $10$. Each initial condition is simulated until $T=5$ with $dt=0.01$, and the same magnitude of Gaussian noise is added. TVRegDiff is used to calculate the derivative, and the first and last $30\%$ percent of data is discarded due to aliasing effect. The testing data is generated using the same process and the ratio of training and testing data is $1:1$. The model and maximum magnitude of noise allowed for each values of $K_m$ is summarized in Table~\ref{TableKm}.  These results suggest that as $K_m$ increases, the maximum noise SINDy-PI can handle also increases.
\begin{table}[h]
    \centering
    \caption{The effect of  $K_m$ on the noise robustness of SINDy-PI.}
    \label{TableKm}
    \begin{tabular}{|c|c|c|c|}
    \hline
    $K_m$ & \begin{tabular}[c]{@{}c@{}}Max Magnitude \\ of Noise\end{tabular} & True Model & Identified Model \\ \hline
    0.01      &     0.001                                                              &  $\dot{x}=-\frac{(0.9x-0.006)}{x+0.01}$          &    $\dot{x}=-\frac{(x-7.6144)(x-5.6092)(x-2.5078)(x-0.0072)}{(x-7.6146)(x-5.6125)(x-2.5092)(x+0.0092)}$                   \\ \hline
    0.1      &      0.01                                                             &   $\dot{x}=-\frac{(0.9x-0.06)}{x+0.1}$         &    $\dot{x}=-\frac{(x-5.4132)(x-0.0681)}{(x-5.4144)(x+0.099)}$                   \\ \hline
    1      &        0.05                                                           &   $\dot{x}=-\frac{(0.9x-0.6)}{x+1}$         &      $\dot{x}=-\frac{(x-4.7143)(x-0.6726)}{(x-4.7159)(x+0.9698)}$                 \\ \hline
    10      &       0.1                                                            &    $\dot{x}=-\frac{(0.9x-6)}{x+10}$        &      $\dot{x}=-\frac{0.99x-6.583}{x+11.07}$                 \\ \hline
    \end{tabular}
\end{table}

\end{document}